\documentclass[runningheads]{llncs}

\usepackage{eccv}

\usepackage[width=122mm,left=12mm,paperwidth=146mm,height=193mm,top=12mm,paperheight=217mm]{geometry} 

\usepackage{eccvabbrv}

\usepackage{graphicx}
\usepackage{booktabs}

\usepackage[accsupp]{axessibility}  

\usepackage[utf8]{inputenc}
\usepackage[ruled,vlined]{algorithm2e} 

\usepackage{multirow}                 

\usepackage{bbding}
\usepackage{pifont}
\usepackage{wasysym}
\usepackage{amssymb}

\usepackage[normalem]{ulem}
\useunder{\uline}{\ul}{}      

\usepackage{arydshln} 

\usepackage[misc]{ifsym}      

\usepackage{hyperref}

\begin{document}

\title{Robust Self-Supervised Cross-Modal Super-Resolution against Real-World Misaligned Observations} 

\titlerunning{RobSelf}  

\author{Xiaoyu Dong\inst{1} \and
Jiahuan Li\inst{1,2} \and
Ziteng Cui\inst{1} \and
Naoto Yokoya\inst{1,2\,\textrm{\Letter}}}

\authorrunning{X.~Dong et al.}

\institute{The University of Tokyo, Japan \and
RIKEN AIP, Japan} 

\maketitle

\begin{figure}[tb]
  \centering
  \includegraphics[width=1\linewidth]{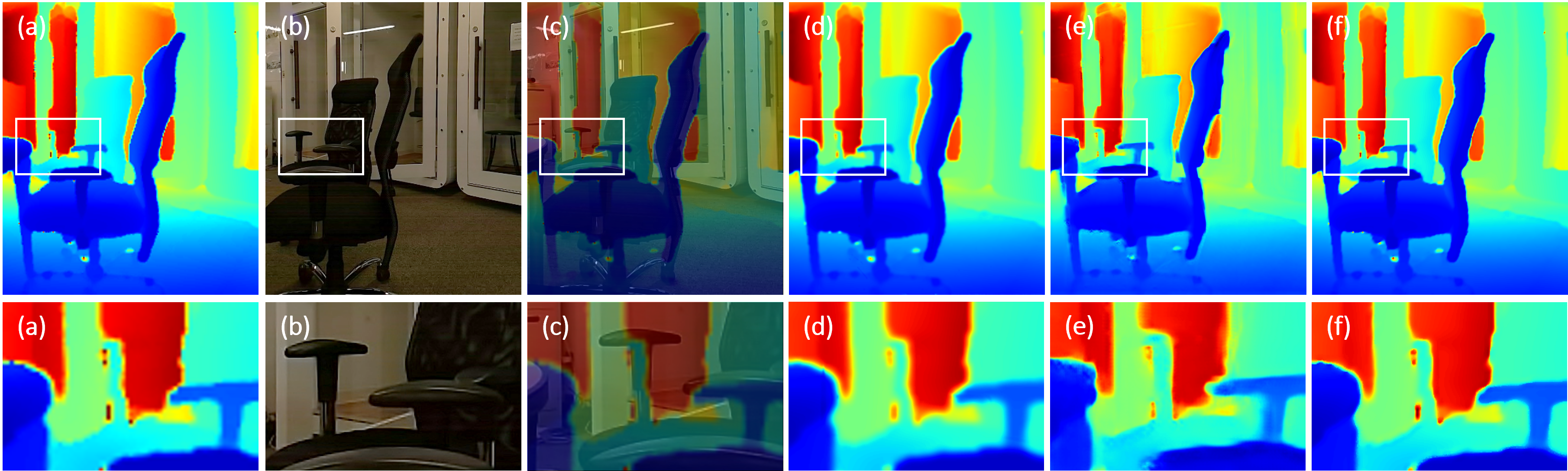} 
  \caption{
  Real-world misaligned RGB-guided depth SR ($\times 4$).
  Our model achieves state-of-the-art performance, \textit{requiring no training data, ground-truth supervision, or pre-alignment.} 
  (a) LR source; (b) HR guide; (c) pre-aligned guide by MINIMA~\cite{MINIMA_2025_CVPR}; (d)~SSGNet~\cite{SSGNet_2023_AAAI} + pre-alignment; (e) SGNet~\cite{SGNet_2024_AAAI} + pre-alignment; 
  (f) RobSelf-Re (Ours).
  }
  \label{fig1}
\end{figure}

\begin{abstract} 
Cross-modal super-resolution (SR) on real-world misaligned data is challenging, as only unlabeled low-resolution (LR) source and high-resolution (HR) guide images with complex spatial misalignment are available. Previous methods either rely on simulated training data or adopt suboptimal alignment strategies that overlook cross-modal dependencies, limiting their practical performance. 
To address these issues, we propose RobSelf, a self-supervised model that jointly optimizes a misalignment-aware feature translator and a content-aware reference filter online. The translator resolves unsupervised cross-modal and cross-resolution alignment via weakly-supervised, misalignment-aware translation, yielding an aligned guide feature. Guided by this feature, the filter performs reference-based discriminative self-enhancement on the source, enabling SR prediction with high resolution and high fidelity. 
Experiments on synthesized data and collected real-world data demonstrate that RobSelf achieves state-of-the-art performance, outperforming existing self-supervised and supervised methods. Moreover, it achieves superior efficiency, being up to 15.3$\times$ faster than prior self-supervised methods. 
\url{https://github.com/palmdong/RobSelf} 
\end{abstract} 


\section{Introduction}

Multi-modal images, \eg, RGB, depth, and near-infrared (NIR), capture complementary properties of objects and environments~\cite{Dong_2024_WACV,OmniSegment_2025_NIPS,DDFM_2023_ICCV,Equi_2024_CVPR}. 
However, non-visible modalities generally suffer from lower spatial resolution than RGB due to sensing and hardware limits, which hinders downstream tasks and necessitates cross-modal super-resolution (SR)~\cite{DCNAS_2025_TPAMI,GraphUnfold_2026_CVPR,DegBins_2026}.

Cross-modal SR enhances a low-resolution (LR) source image using structural cues from a high-resolution (HR) guide image of another modality. 
Modern methods fall into supervised and self-supervised categories. 
Supervised methods rely on large-scale domain-specific training data and ground truth~\cite{DORNet_2025_CVPR,DCNAS_2025_TPAMI,CP2D_2025_AAAI,SGNet_2024_AAAI,GraphUnfold_2026_CVPR,SPFNet_2026_IJCV}. 
This reliance hinders their practical generalization, as building such datasets is costly and labor-intensive. 
Self-supervised methods require neither training data nor ground truth, but instead optimize online on each test pair, making them more data-efficient and generalizable~\cite{SSGNet_2023_AAAI,MMSR_ECCV_2022,CMSR_2021_CVPR,P2P_2019_ICCV}.


While promising progress has been achieved, most existing methods assume that the source and guide images are well-aligned ~\cite{DORNet_2025_CVPR,DCNAS_2025_TPAMI,CP2D_2025_AAAI,SGNet_2024_AAAI,SSGNet_2023_AAAI,MMSR_ECCV_2022,P2P_2019_ICCV,GraphUnfold_2026_CVPR,SPFNet_2026_IJCV}. 
In real-world scenarios, however, spatial misalignment is inevitable in multi-modal images
due to inherent cross-sensor discrepancies~\cite{DEPTHOR_2025_ICCV,Studio_2025_CVPR} (\eg, differences in lens distortion, field of view, and physical position) and environmental factors, such as platform-induced viewpoint variation~\cite{PixelNIR_2025_CVPR,LuGS++_2026} and object motion over time~\cite{ToFGaussian_2025_CVPR,Video_2026_AAAI}. 
Although several methods consider misalignment, they either rely on simulated training data~\cite{MOMNet_2026} or adopt suboptimal alignment strategies~\cite{CMSR_2021_CVPR,UGSR_2021_TIP}, limiting their practical performance.

We identify two difficulties in cross-modal SR on \textit{real-world} misaligned data:
(\uppercase\expandafter{\romannumeral1}) The scarcity of training data and the absence of SR ground truth for the source~\cite{P2P_2019_ICCV,MMSR_ECCV_2022};
(\uppercase\expandafter{\romannumeral2})~Complex misalignments across modalities and resolutions, along with the lack of alignment ground truth for the guide~\cite{CrossHomo_2024_PAMI,SCPNet_2024_ECCV}.
These hinder the development of reliable supervised methods.
One solution is to perform pre-alignment before SR.
However, such a two-stage pipeline may not generalize effectively to real-world data with complex misalignment and resolution gap (\cref{fig1}). 
Overall, robust self-supervised cross-modal SR for real-world misaligned data is crucial yet remains an open challenge. 

\begin{figure}[!t]
\centering
\includegraphics[width=1\linewidth]{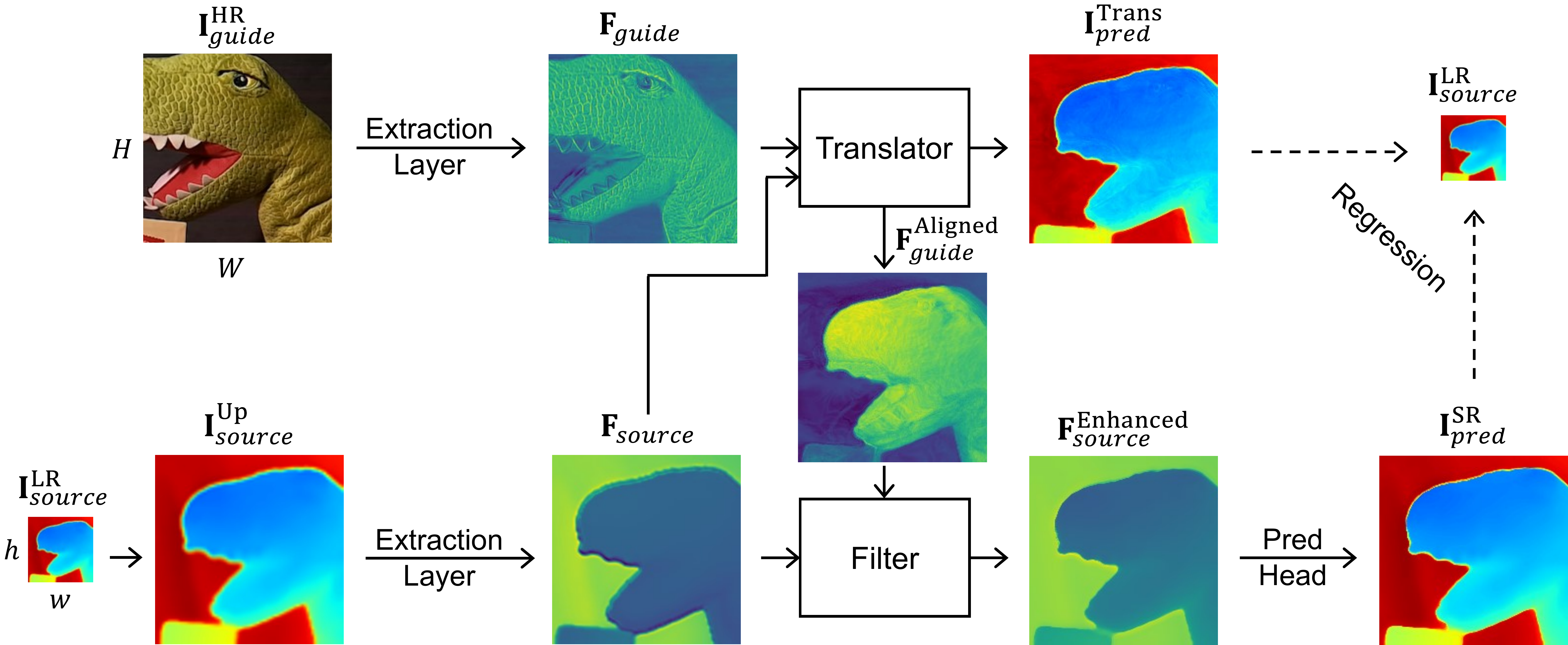} 
\caption{
\textbf{RobSelf} is solely supervised by the LR source. 
Within this framework, the translator is optimized under weak supervision to map $\mathbf{F}_{guide}$ into $\mathbf{I}^{\rm Trans}_{pred}$ that mimics the source,
while yielding $\mathbf{F}^{\rm Aligned}_{guide}$. 
Guided by $\mathbf{F}^{\rm Aligned}_{guide}$, the filter performs reference-based discriminative self-enhancement on $\mathbf{F}_{source}$ to obtain $\mathbf{F}^{\rm Enhanced}_{source}$, from which $\mathbf{I}^{\rm SR}_{pred}$ is generated. 
RobSelf has {\ul two variants} depending on the the translator design. 
}
\label{fig2:model}
\end{figure}

To address these issues, we propose RobSelf (\cref{fig2:model}), a self-supervised model that resolves unsupervised cross-modal and cross-resolution alignment via a formulation of joint weakly-supervised, misalignment-aware translation. 
RobSelf features a misalignment-aware feature translator (\cref{fig3:translator}) and a content-aware reference filter (\cref{fig4:filter}).
The translator warps and translates the guide feature to mimic the source modality, while yielding an aligned guide feature with inherent redundancy.
Guided by this feature, the filter learns content-aware kernels for reference-based discriminative self-enhancement on the source, enabling effective enhancement while mitigating redundancy effects.
To evaluate RobSelf, we collected real-world RGB-depth and RGB-NIR data with inherent cross-sensor misalignment, random viewpoint variation, and random object motion. 
Experiments demonstrate that RobSelf achieves HR and high-fidelity predictions on complex real-world misaligned data, even without training data, ground-truth supervision, or pre-alignment (\cref{fig1}).
Moreover, it achieves superior efficiency, being up to 15.3$\times$ faster than prior self-supervised methods~\cite{P2P_2019_ICCV,SSGNet_2023_AAAI,MMSR_ECCV_2022}.

Contributions: 
(\uppercase\expandafter{\romannumeral1}) We address the open challenge of robust self-supervised cross-modal SR on real-world misaligned data by proposing RobSelf. 
(\uppercase\expandafter{\romannumeral2}) We propose a joint weakly-supervised, misalignment-aware translation formulation that effectively handles diverse misalignment conditions and even boundary scenarios with missing guide structures. 
(\uppercase\expandafter{\romannumeral3}) Our reference-based discriminative self-enhancement strategy enables faithful source enhancement without additional guide processing. 
(\uppercase\expandafter{\romannumeral4}) Experiments on synthesized depth SR, real-world depth SR, and real-world NIR SR demonstrate state-of-the-art performance, robustness, and generalizability.

\section{Related Work} 

\subsection{Cross-Modal SR}

Modern learning-based cross-modal SR methods fall into supervised and self-supervised categories. 
Supervised methods rely on large-scale domain-specific training data and ground truth~\cite{DORNet_2025_CVPR,DCNAS_2025_TPAMI,CP2D_2025_AAAI,DCTNet_2022_CVPR,FDSR_2021_CVPR,Spherical_2023_ICCV,ULB17-VT_2018,DuCos_2025_ICCV,GraphUnfold_2026_CVPR,SPFNet_2026_IJCV,Yan_2022_TNNLS}. 
Self-supervised methods~\cite{SSGNet_2023_AAAI,MMSR_ECCV_2022,CMSR_2021_CVPR,P2P_2019_ICCV,CDFDSR_2025_IF} are more data-efficient and generalizable, requiring no training data or ground-truth supervision. 
However, most methods overlook misalignment issues, assuming spatially well-aligned inputs.
For example, SFG~\cite{SFG_2023_AAAI} 
addresses structural distortion and edge noise in depth SR under aligned RGB guidance. 
To address misalignment, MOMNet~\cite{MOMNet_2026}, a concurrent supervised method, develops multi-order feature matching modules using simulated training data, which do not fully capture real-world misalignments. 
CMSR~\cite{CMSR_2021_CVPR} addresses misalignment via a self-supervised framework built on pseudo-paired data. 
However, it relies on off-the-shelf RGB deformation layers and overlooks cross-modal dependencies, resulting in suboptimal performance. 

In contrast:
(\uppercase\expandafter{\romannumeral1}) We address \textit{complex real-world misalignments} arising from sensor discrepancies and environmental factors. 
(\uppercase\expandafter{\romannumeral2}) Instead of relying on modules designed solely for image or feature alignment~\cite{CMSR_2021_CVPR,MOMNet_2026,UGSR_2021_TIP}, we propose a misalignment-aware \textit{feature translator} that, driven by its weakly-supervised translation objective, effectively handles \textit{diverse misalignments and missing guide structures} within a self-supervised SR framework. 

Misalignment also arises in guided hyperspectral image (HSI) SR~\cite{UnalignedHSI_2025_IJCV,UnalignedHSI_2024_TNNLS,UnalignedHSI_2024_AAAI}, pan-sharpening~\cite{PAN_2025_ICCV}, fusion~\cite{MulFS-CAP_2025_PAMI,MURF_2023_PAMI,Fusion_2025_TIP,Video_2025_NIPS}, MRI reconstruction~\cite{Reconstruct_2022_TMI}, and depth refinement~\cite{Refine_2019_ICCV}. 
Unlike these methods that rely on training data and focus on image- or feature-level registration, 
our online self-supervised model with a joint translation objective achieves better robustness 
(\cref{subsec:real_rgb_nir}). 


\subsection{Multi-Modal Image Registration}
\label{subsec:mm_registrtaion}

Multi-modal image registration aims to align images of different modalities into a common coordinate space. 
Modern
data-driven methods estimate transformation parameters or deformation fields in supervised~\cite{MINIMA_2025_CVPR,SDME_2024_ICML,AgnosticMedical_2024_CVPR,CrossHomo_2024_PAMI,Recurrent_2023_CVPR,Indescribable_2023_CVPR,Lucas_2021_CVPR} and unsupervised~\cite{Alternating_2024_NIPS,SCPNet_2024_ECCV,RS_2023_TGRS,InvertTrans_2024_ECCV,Trans_2020_CVPR} paradigms. 
For example, Ren~\etal~\cite{MINIMA_2025_CVPR} proposed a supervised model, MINIMA, for unified cross-modal matching.
Deng~\etal~\cite{CrossHomo_2024_PAMI} developed a supervised model for cross-modal, cross-resolution homography estimation.
Recently, Song \etal~\cite{Alternating_2024_NIPS} developed an unsupervised registration framework via alternating optimization.
However, data-driven methods rely on training data and may not generalize well to complex in-the-wild scenarios. 

For our task, training data are scarce and neither SR nor alignment ground truth is available. To address this challenge, we cast unsupervised cross-modal and cross-resolution alignment as a weakly-supervised, misalignment-aware translation subtask 
within RobSelf for \textit{joint online optimization}. This enables robustness and generalizability across diverse data and misalignments. 


\subsection{Learnable Image Filtering} 

Learnable image filtering learns filter rules or weights from data, enabling adaptive content modeling, suppression, and extraction~\cite{Bandpass_2020_CVPR,LowPassDeblur_2023_ICCV}. 
In high-level vision, learnable filters such as tree~\cite{LTF_2019_NIPS,LTFv2_2020_NIPS} and non-local~\cite{NL_2018_CVPR,NLCRF_2022_CVPR} filters have proven effective for modeling dependencies and preserving structures. 
In low-level vision, non-local filters are effective in exploiting self-similarity priors to preserve structures~\cite{Pansharpening_2024_CVPR}, remove noise~\cite{NLRN_2018_NIPS}, and recover details~\cite{Mei_2021_CVPR}. 
In cross-modal SR, degradation-aware filters~\cite{DORNet_2025_CVPR}, cross-domain adaptive filters~\cite{MMSR_ECCV_2022}, and deformable filters~\cite{DKN_2021_IJCV} have been introduced to facilitate multi-modal feature processing and fusion.
These filters are, however, designed for well-aligned data and may fail to transfer guidance information reliably when misalignments occur. 

We further observe that the guide, even after alignment (translation), contains both structures essential for source enhancement and redundant content due to modality discrepancies. 
Therefore, \textit{rather than filtering or fusing the guide}, 
we design a filter that learns content-aware kernels to perform reference-based discriminative self-enhancement on the source. 
This enables HR and faithful SR predictions, free from the effects of redundant guide content.





\section{RobSelf}

\subsection{Overview} 

As illustrated in \cref{fig2:model},
given an input image pair, the LR source $\mathbf{I}^{\rm LR}_{source}\in\mathbb{R}^{h\times w\times \phi}$ is first bilinearly upsampled to match the spatial size of the HR guide $\mathbf{I}^{\rm HR}_{guide}\in\mathbb{R}^{H\times W\times \psi}$. 
Two extraction layers then produce the source and guide features, $\mathbf{F}_{source},\mathbf{F}_{guide}\in\mathbb{R}^{C\times H\times W}$. 
$\phi$, $\psi$, and $C$ denote channel dimensions.
The translator maps $\mathbf{F}_{guide}$ to an HR prediction $\mathbf{I}^{\rm Trans}_{pred}$, while yielding the aligned guide feature $\mathbf{F}^{\rm Aligned}_{guide}$.
Guided by $\mathbf{F}^{\rm Aligned}_{guide}$, the filter enhances $\mathbf{F}_{source}$ to obtain $\mathbf{F}^{\rm Enhanced}_{source}$, from which the SR prediction $\mathbf{I}^{\rm SR}_{pred}$ is generated by a prediction head.
Both $\mathbf{I}^{\rm SR}_{pred}$ and $\mathbf{I}^{\rm Trans}_{pred}$ are supervised by regression to the LR source via a consistency loss~\cite{MMSR_ECCV_2022}:
\begin{equation}
  \mathcal{L} = \mathcal{L}_{sr} + \lambda\mathcal{L}_{trans} 
  = \big\Vert f_{down}(\mathbf{I}^{\rm SR}_{pred}) - \mathbf{I}_{source}^{\rm LR} \big\Vert_1 + 
  \lambda \big\Vert f_{down}(\mathbf{I}^{\rm Trans}_{pred}) - \mathbf{I}_{source}^{\rm LR} \big\Vert_1,
  \label{eq_loss}
\end{equation}   
where $f_{down}$ denotes average pooling and $\lambda$ is a weighting factor. 
Network details are provided in the supplementary material.


\begin{figure}[!b]
\centering
\includegraphics[width=1\linewidth]{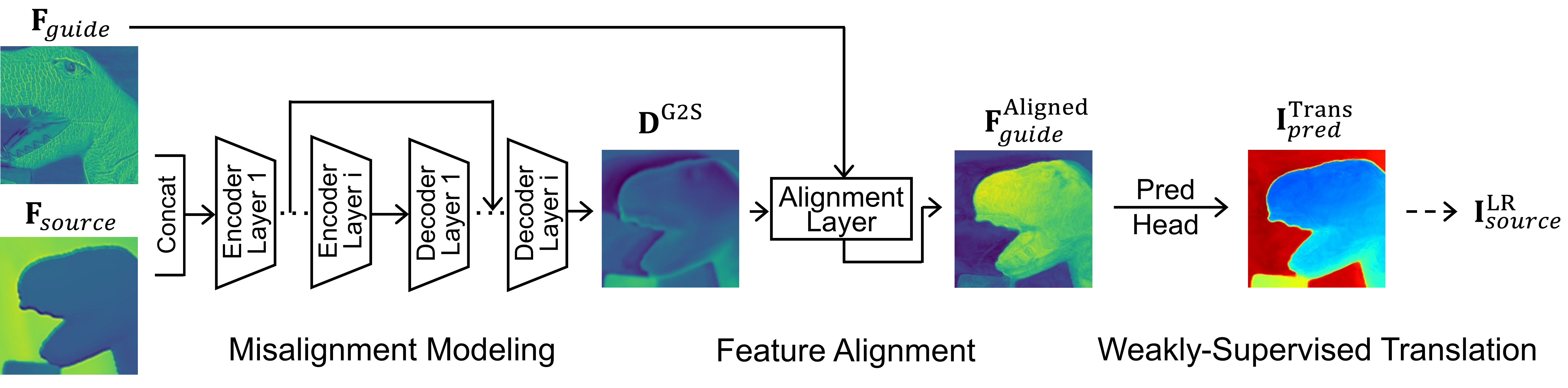} 
\caption{
Misalignment-aware feature translator.
Each encoder layer downsamples by $\times 2$; each decoder layer upsamples by $\times 2$.
}
\label{fig3:translator}
\end{figure}

\subsection{Misalignment-Aware Feature Translator} 
\label{subsec:translator}

As shown in (\cref{fig3:translator}),
driven by the translation objective mimicking the source modality, 
the translator yields $\mathbf{F}^{\rm Aligned}_{guide}$, which is aligned with the source and contains both essential structures and redundant content. 

\textbf{Misalignment Modeling.} 
The translator first derives a cross-modal deformation field $\mathbf{D}^\text{G2S}$ that models the misalignment from $\mathbf{F}_{guide}$ to $\mathbf{F}_{source}$ using a multi-level ($i$-level) estimator. 
With progressive downsampling and upsampling, this estimator captures misalignments across multiple spatial scales.
When $i=4$, for example, the input features are downsampled by 16 after the encoder, enabling misalignment modeling at the regional level.
$\mathbf{D}^\text{G2S}$ is an $H \times W$ matrix of $N$-dimensional offset vectors. 

\textbf{Feature Alignment.}
The alignment layer warps $\mathbf{F}_{guide}$ according to $\mathbf{D}^\text{G2S}$, yielding $\mathbf{F}^{\rm Aligned}_{guide}$.
It is implemented by a deformable convolution (conv)~\cite{DCN_2017_ICCV} or a simple spatial resampling operation, resulting in {\ul two model variants: RobSelf-De and RobSelf-Re}.
Formally, at location $p$:
\begin{equation}
  \mathbf{F}^{\rm Aligned}_{guide}(p) =
  \begin{cases}
    \displaystyle \sum_{i=1}^{k^2} w_i
    \mathbf{F}_{guide}\big(p + p_i + \Delta p_i\big), 
    & \text{if RobSelf-De}, \\[12pt]
    \mathbf{F}_{guide}\big(p + \Delta p\big), 
    & \text{if RobSelf-Re},
  \end{cases}
  \label{eq_aligned_guide}
\end{equation}
where $p_i$ denotes the $i$-th relative location in the $k\times k$ conv kernel centered at $p$, $w_i$ is the learned kernel weight, 
and the sampling offsets $\Delta p_i=(\Delta x_i, \Delta y_i)$ 
and $\Delta p=(\Delta x, \Delta y)$ are learned from $\mathbf{D}^\text{G2S}$. 
Accordingly, the dimension $N$ of $\mathbf{D}^\text{G2S}$ is $2k^2$ for RobSelf-De and $2$ for RobSelf-Re. 

\textbf{Translation Objective.} 
Through a prediction head, the translator finally generates an HR prediction $\mathbf{I}^{\rm Trans}_{pred}$, weakly supervised by $\mathbf{I}^{\rm LR}_{source}$ to drive effective feature alignment. 
Such a joint weakly-supervised, misalignment-aware translation formulation enables robustness and generalizability across diverse data and misalignments, as well as the ability to handle missing guide structures even within a self-supervised SR framework 
(\cref{subsec:interesting}).


\begin{figure}[!b]
\centering
\includegraphics[width=0.85\linewidth]{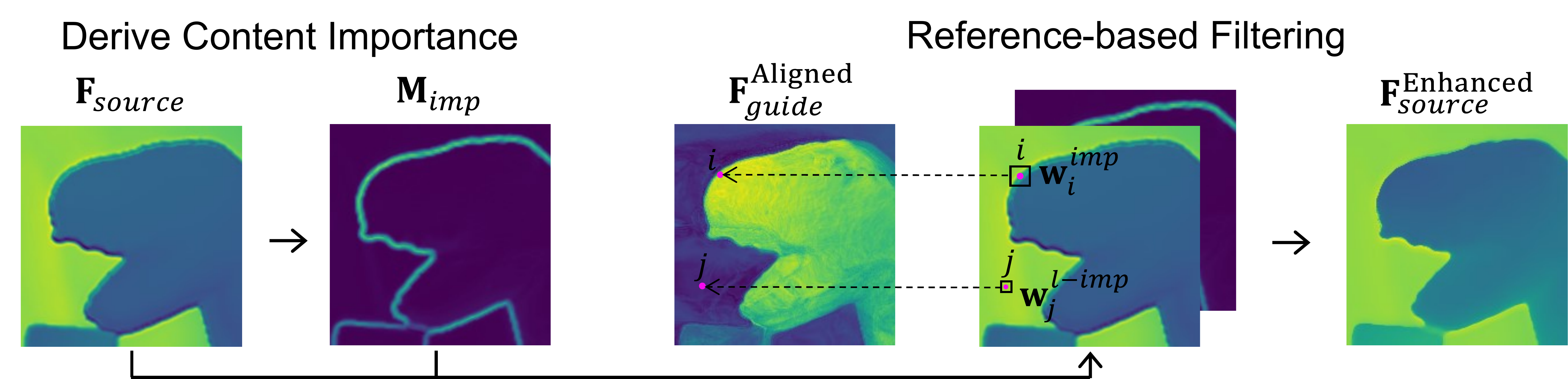} 
\caption{
Content-aware reference filter.
}
\label{fig4:filter}
\end{figure}

\subsection{Content-Aware Reference Filter} 
\label{subsec:filter}

To better exploit essential guidance and mitigate redundancy effects,
we design a filter (\cref{fig4:filter}) that learns content-aware kernels to enhance $\mathbf{F}_{source}$ using its own pixels, with $\mathbf{F}^{\rm Aligned}_{guide}$ only as a reference for weight determination—essentially reference-based discriminative self-enhancement. 

\textbf{Content-Aware Design.} 
We first derive an importance map $\mathbf{M}_{imp}\in\mathbb{R}^{H\times W}$:
\begin{equation}
\mathbf{M}_{imp} = \frac{1}{C} \sum_{c=1}^{C} \big\|\,\nabla (\mathbf{F}_{source}^c)\,\big\|_{2}, 
\label{eq_importance}
\end{equation}
where $\nabla$ denotes spatial gradient operator, $c$ is channel index, and $C$ is number of channels. 
The motivation is: 
high-gradient source content (\eg, edges and textures) is important, corresponding to essential guide structures; 
low-gradient source content (\eg, smooth regions) is less important,
corresponding to redundant guide content. 
Based on content importance, we adopt large and small kernels to discriminatively enhance the source pixels. 

\textbf{Reference-based Filtering.} 
A threshold is computed as: 
\begin{equation}
\tau = \eta\,\mathbb{E}\!\left[\mathbf{M}_{imp}\right], 
\label{eq_threshold}
\end{equation} 
where $\eta$ is a scaling factor. 
For important pixels $\mathbf{F}_{source}(i)$ with $\mathbf{M}_{imp}(i)>\tau$, a large kernel $\mathbf{w}_i^{imp}\in\mathbb{R}^{m\times m}$ aggregates more neighbors for stronger self-enhancement under essential guidance. 
For less important pixels $\mathbf{F}_{source}(j)$ with $\mathbf{M}_{imp}(j)<\tau$, a small kernel $\mathbf{w}_j^{l\text{-}imp}\in\mathbb{R}^{n\times n}$ performs lightweight self-update under redundant guidance. 
We learn correlation-based kernel weights~\cite{MMSR_ECCV_2022},
proven effective in inducing pixels to emulate the properties of cross-domain references.
The weights of $\mathbf{w}_i^{imp}$, for example, are determined by the correlation between the source pixels within an $m\times m$ neighborhood $\mathcal{N}_i$ centered at $\mathbf{F}_{source}(i)$ and the reference pixel $\mathbf{F}^{\rm Aligned}_{guide}(i)$ 
with inherent HR properties: 
\begin{equation}
w_{i,l}^{imp} =
\frac{\exp\!\big(\,\mathbf{F}_{source}(l)^{\top}\,\mathbf{F}^{\rm Aligned}_{guide}(i)\big)}
{\sum\limits_{l'\in\mathcal{N}_i}
\exp\!\Big(\,\mathbf{F}_{source}(l')^{\top}\,\mathbf{F}^{\rm Aligned}_{guide}(i)\Big)},
\label{eq_single_weight} 
\end{equation}
where $w_{i,l}^{imp}$ is the weight for a specific neighbor $\mathbf{F}_{source}(l)$ of $\mathbf{F}_{source}(i)$, and $l'$ indexes pixels in $\mathcal{N}_i$ for normalization.
The self-enhanced pixel is obtained as: 
\begin{equation}
\mathbf{F}^{\rm Enhanced}_{source}(i) = \sum_{l\in \mathcal{N}_i} w_{i,l}^{imp} \mathbf{F}_{source}(l). 
\label{eq_enhanced_pixel} 
\end{equation}

Such reference-based discriminative self-enhancement 
requires no additional guide processing and enables $\mathbf{F}^{\rm Enhanced}_{source}$ and $\mathbf{I}^{\rm SR}_{pred}$ with high resolution and high fidelity, free from the effects of redundant content in $\mathbf{F}^{\rm Aligned}_{guide}$. 



\section{Real-World Misaligned Data} 

\begin{figure}[!t]
\centering
\begin{subfigure}{0.485\linewidth} 
\includegraphics[width=0.49\linewidth]{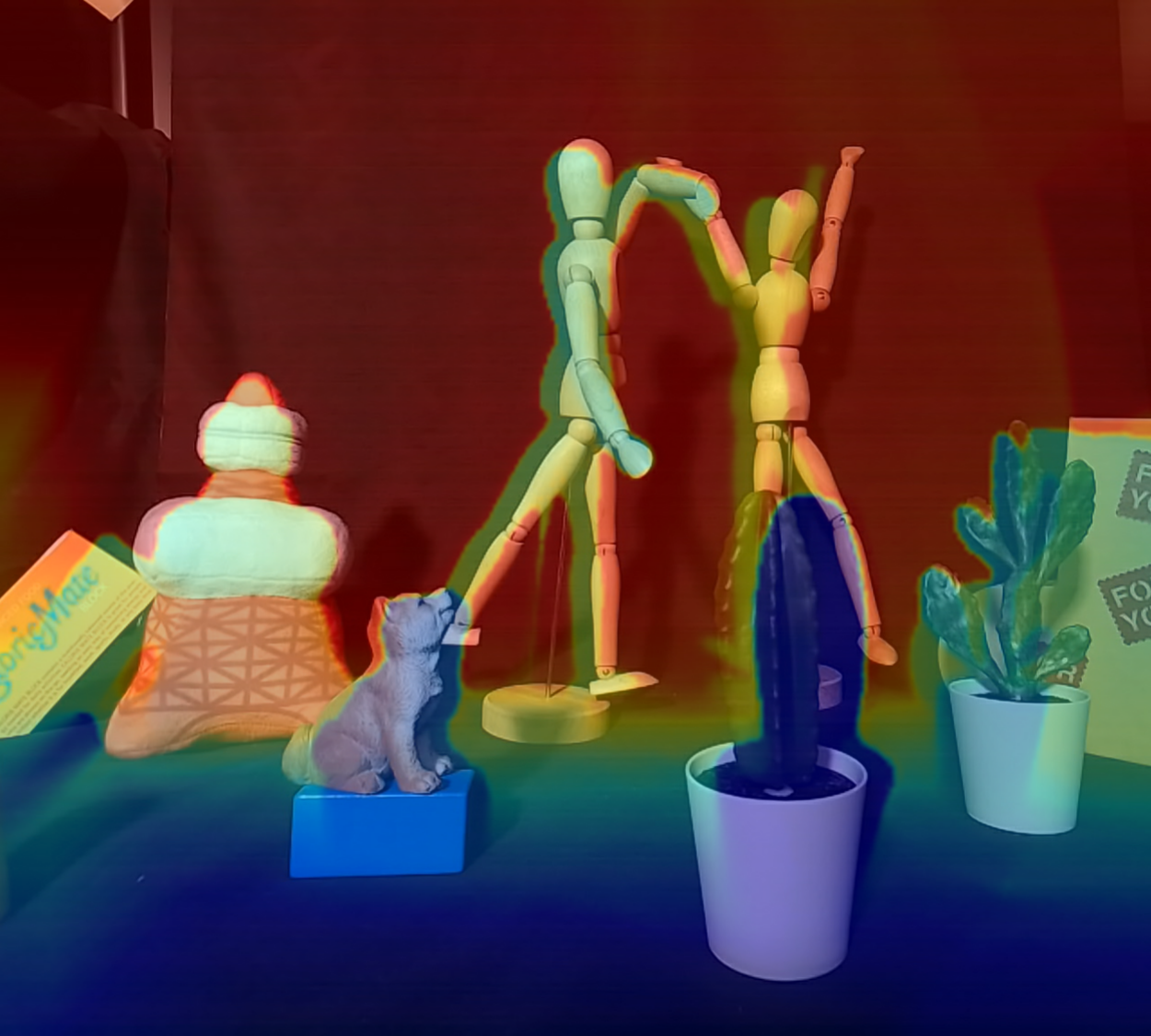} 
\includegraphics[width=0.49\linewidth]{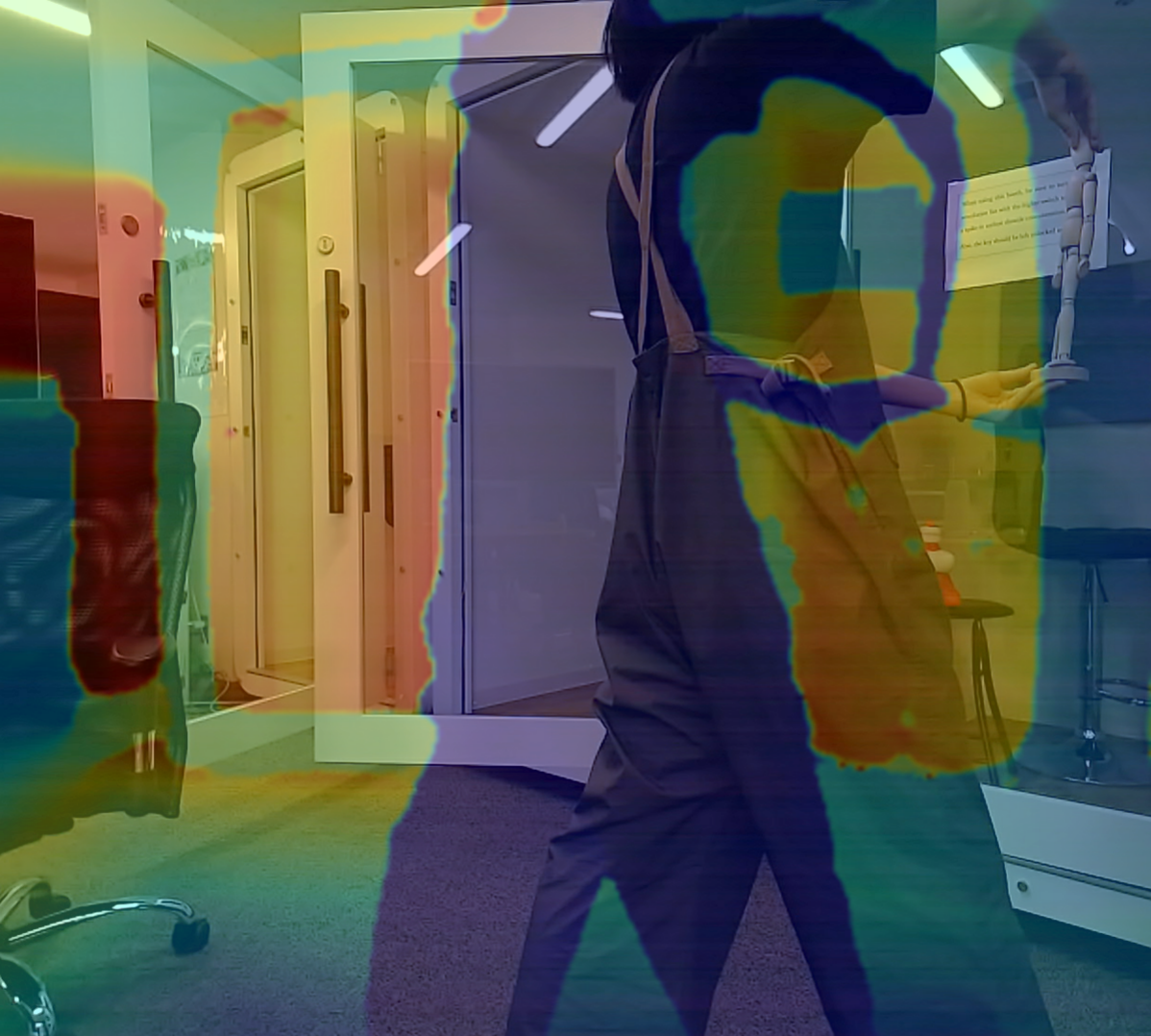}
\subcaption{\textbf{RGB-Depth}. Left: Inherent cross-sensor misalignment. Right: Inherent cross-sensor misalignment and random viewpoint variation.}
\end{subfigure}
{~}
\begin{subfigure}{0.485\linewidth}
\includegraphics[width=0.49\linewidth]{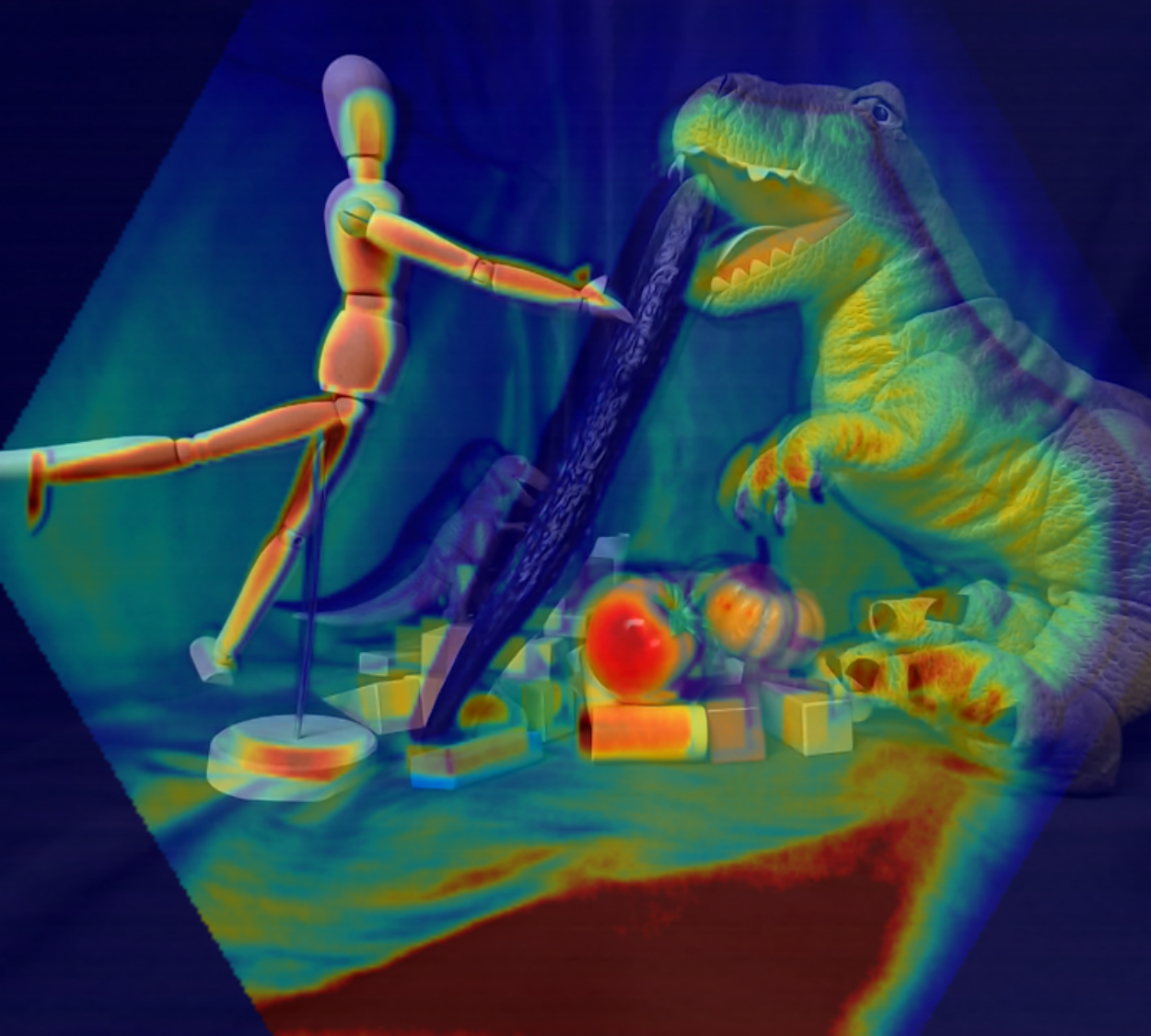} 
\includegraphics[width=0.49\linewidth]{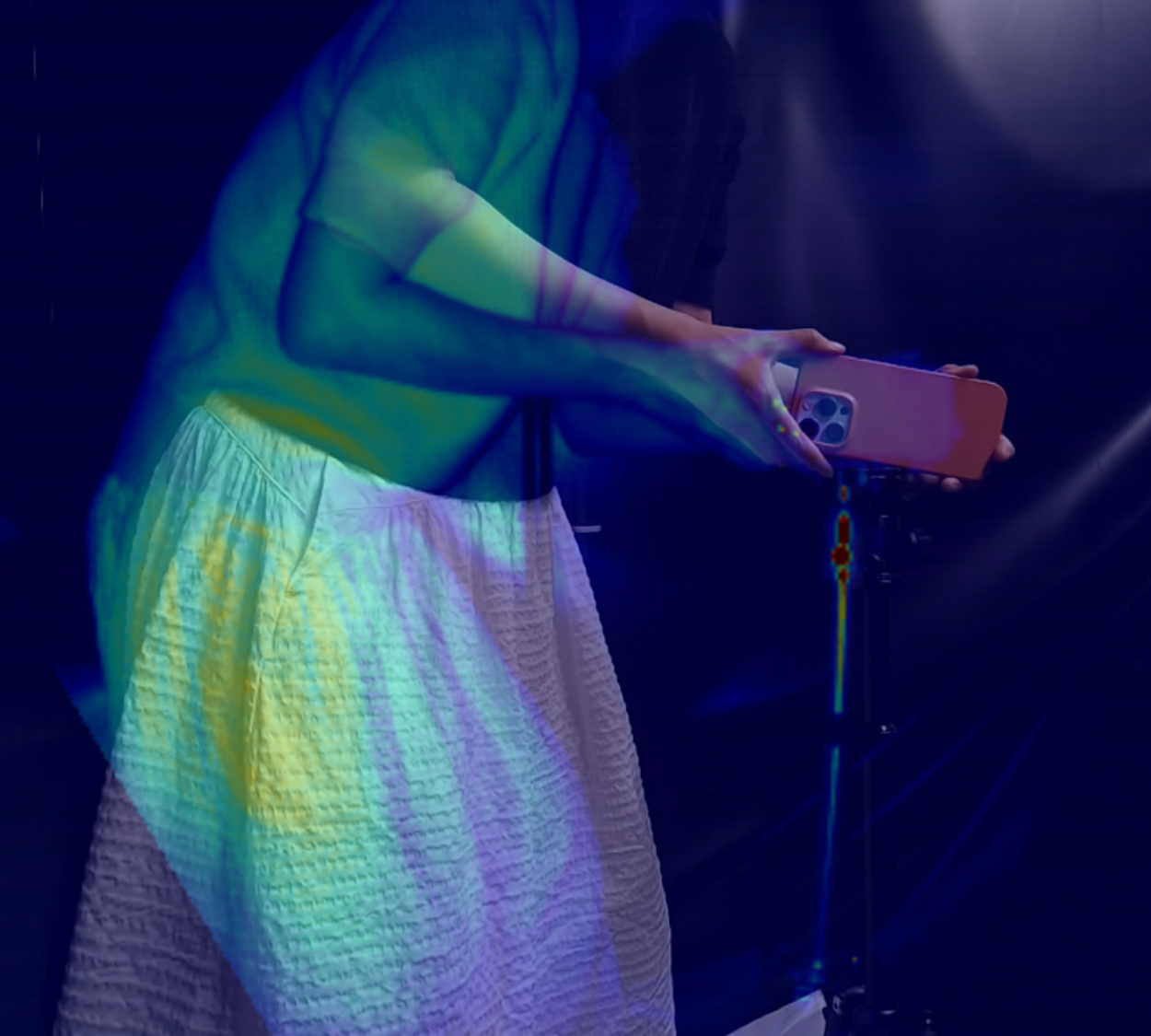} 
\subcaption{\textbf{RGB-NIR}. Left: Inherent cross-sensor misalignment. Right: Inherent cross-sensor misalignment and random object motion.}
\end{subfigure}
\caption{
Examples from our collected real-world misaligned data.
LR sources are overlaid on HR guides for better visualization of misalignments.
}
\label{fig5:RealMisSR_example}
\end{figure}

To evaluate RobSelf in real-world scenarios, we collected data with two modality combinations, three types of misalignments, real LR data, and diverse scenes and objects (\cref{fig5:RealMisSR_example}). 
Following~\cite{UGSR_2021_TIP,UnalignedHSI_2024_TNNLS}, the data were captured using a single multi-modal device (Azure Kinect DK in our case). 
To avoid bias toward fixed cross-sensor misalignment, we introduced random viewpoint variations and object motions during data acquisition.
Collection procedures and data comparisons are provided in the supplementary material. 

The RGB-Depth subset provides 52 groups of simple cases with inherent cross-sensor misalignment (caused by factors such as sensors' lens distortion, fields of view, and physical positions), and 60 groups of complex cases with inherent cross-sensor misalignment and random viewpoint variation. 
Each group contains a raw LR depth, a filled LR depth, a $\times 2$ HR depth, a $\times 2$ HR RGB, and a $\times 4$ HR RGB. The data cover small-scale toy scenes and large-scale indoor scenes, with objects such as toys, desks, chairs, and other common indoor items. 
The RGB-NIR subset provides 50 groups of simple cases with inherent cross-sensor misalignment, and 30 groups of complex cases with inherent cross-sensor misalignment and random object motion. 
Each group contains an LR NIR, a $\times 2$ HR NIR, a $\times 2$ HR RGB, and a $\times 4$ HR RGB. The data span small-scale toy scenes and medium-scale human-centric scenes, with objects such as real plants, artificial plants, toys, and humans. 
For both subsets, there is no ground truth for $\times 4$ SR. 
The resolutions are: LR—288$\times$320, $\times 2$ HR—576$\times$640, and $\times 4$ HR—1152$\times$1280. 

\section{Experiments}
\label{sec:experiments}

\subsection{Settings}
\label{subsec:settings}

\textbf{Tasks.} 
We conduct three tasks: 
(\uppercase\expandafter{\romannumeral1})~Synthesized misaligned RGB-guided depth SR ($\times 4$ and $\times 8$); (\uppercase\expandafter{\romannumeral2})~Real-world misaligned RGB-guided depth SR ($\times 2$ and $\times 4$); (\uppercase\expandafter{\romannumeral3})~Real-world misaligned RGB-guided NIR SR ($\times 2$ and $\times 4$). 
For task~\uppercase\expandafter{\romannumeral1}, we use the Middlebury dataset~\cite{Scharstein_2007_CVPR,Hirschmuller_2007_CVPR}, with misalignments synthesized via random translation, rotation, and perspective shifts, and LR data generated following~\cite{DADA_2023_CVPR}. 
For tasks \uppercase\expandafter{\romannumeral2} and \uppercase\expandafter{\romannumeral3}, we use the complex cases from our collected data: RGB-depth with inherent cross-sensor misalignment and random viewpoint variation, and RGB-NIR with inherent cross-sensor misalignment and random object motion. 
Pre-alignment was performed for methods lacking alignment strategies using the pre-trained MINIMA model~\cite{MINIMA_2025_CVPR}.

\textbf{Implementation.}   
Our model is implemented in PyTorch on an NVIDIA A100 GPU.
To ensure fairness, optimization is conducted per image pair for 1000 iterations following~\cite{MMSR_ECCV_2022}, without data augmentation or pre-training. 
We adopt the Adam optimizer~\cite{Adam_2015_ICLR} with an initial learning rate of 0.001, decayed by 0.9998 every 5 iterations. 
The loss weight $\lambda$ is set to 1.
The estimator level $i$, filter threshold scaling factor $\eta$, and kernel sizes $\{m,n\}$ slightly vary with tasks. 
Analyses are provided in the supplementary material.
RMSE, DSS (depth-specific, no-reference)~\cite{DSS_2020_SPL}, or NIQE (no-reference)~\cite{NIQE_2015_TIP} is used for evaluation, depending on the availability of ground truth.


\begin{table}[t]
  \caption{
  Synthesized misaligned RGB-guided depth SR.
  For methods lacking alignment strategies, results without and with pre-alignment are reported (wo/w).
  $^*$Params vary with SR factors ($\times4$/$\times8$) due to level $i$.
  The \textbf{best} and {\ul second best} are highlighted.
  } 
  \label{tab1_syn_rgb_depth}
  \centering
  \scalebox{0.62}{
  \begin{tabular}{l|cccccccc|cc}
    \toprule
    & P2P~\cite{P2P_2019_ICCV} & CMSR~\cite{CMSR_2021_CVPR} & DCTNet~\cite{DCTNet_2022_CVPR} & MMSR~\cite{MMSR_ECCV_2022} & SSGNet~\cite{SSGNet_2023_AAAI} & SGNet~\cite{SGNet_2024_AAAI} & C2PD~\cite{CP2D_2025_AAAI} & DORNet~\cite{DORNet_2025_CVPR} & RobSelf-Re & RobSelf-De \\ \hline
    Self-Supervised & $\checkmark$ & $\checkmark$ & \ding{55} & $\checkmark$ & $\checkmark$ & \ding{55} & \ding{55} & \ding{55} & $\checkmark$ & $\checkmark$ \\ 
    Pre-Alignment & $\checkmark$ & \ding{55} & $\checkmark$ & $\checkmark$ & $\checkmark$ & $\checkmark$ & $\checkmark$ & $\checkmark$ & \ding{55} & \ding{55} \\
    Params (M) & 0.2 & 1.18 & 0.48 & 0.25 & 0.31 & 36.39 & 65.06 & 3.05 & 0.76/0.87$^*$ & 0.81/0.92$^*$ \\ \hline
    $\times4$ (RMSE$\downarrow$) & 2.81/2.33 & 1.91 & 2.19/2.16 & 2.22/1.88 & 2.20/1.92 & 2.44/2.33 & 2.12/2.06 & 2.57/2.46 & {\ul 1.52} & \textbf{1.43} \\
    $\times8$ (RMSE$\downarrow$) & 3.33/2.91 & 3.03 & 3.34/3.33 & 2.99/2.79 & 2.99/2.83 & 3.48/3.34 & 3.27/3.15 & 3.58/3.54 & {\ul 2.57} & \textbf{2.49} \\
    \bottomrule
  \end{tabular}
  } 
\end{table}

\begin{figure}[!t]
\centering
\begin{subfigure}{0.19\linewidth} 
\includegraphics[width=1\linewidth]{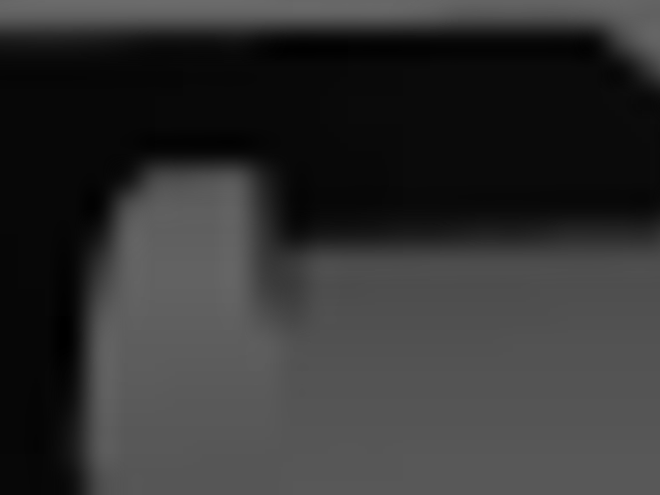}
\subcaption*{LR Source}
\end{subfigure}
\begin{subfigure}{0.19\linewidth}
\includegraphics[width=1\linewidth]{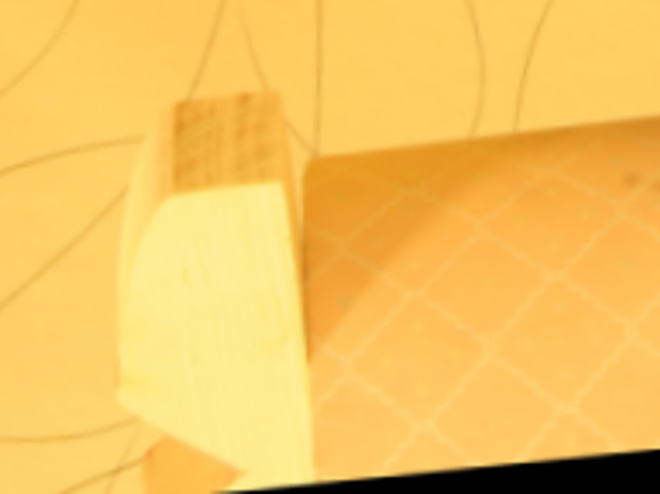}
\subcaption*{HR Guide}
\end{subfigure}
\begin{subfigure}{0.19\linewidth}
\includegraphics[width=1\linewidth]{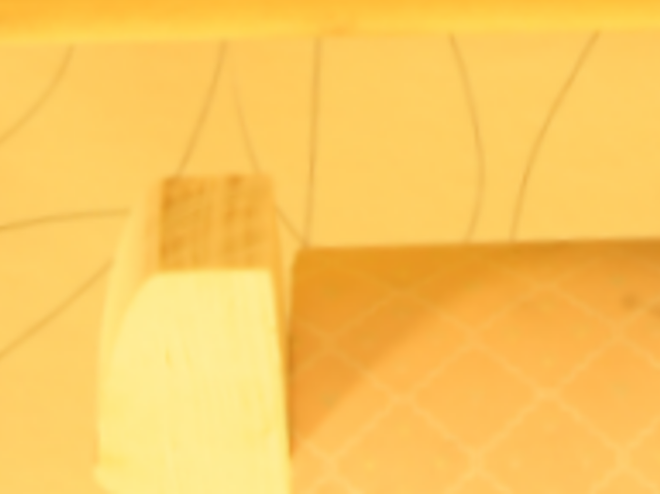}
\subcaption*{Pre-Aligned Guide}
\end{subfigure}
\begin{subfigure}{0.19\linewidth}
\includegraphics[width=1\linewidth]{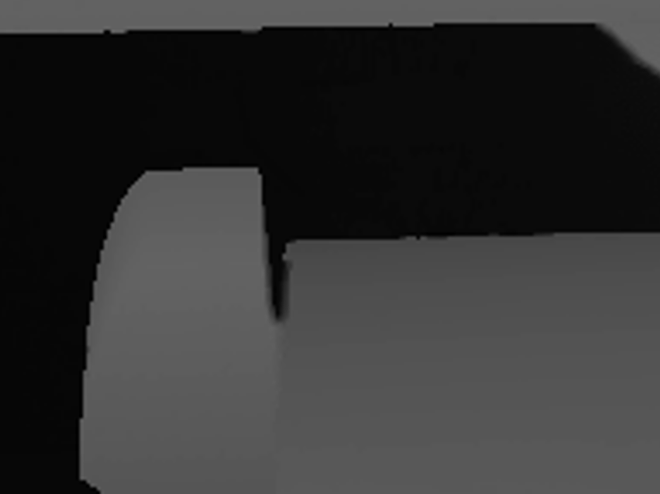}
\subcaption*{Ground Truth}
\end{subfigure}
\begin{subfigure}{0.19\linewidth}
\includegraphics[width=1\linewidth]{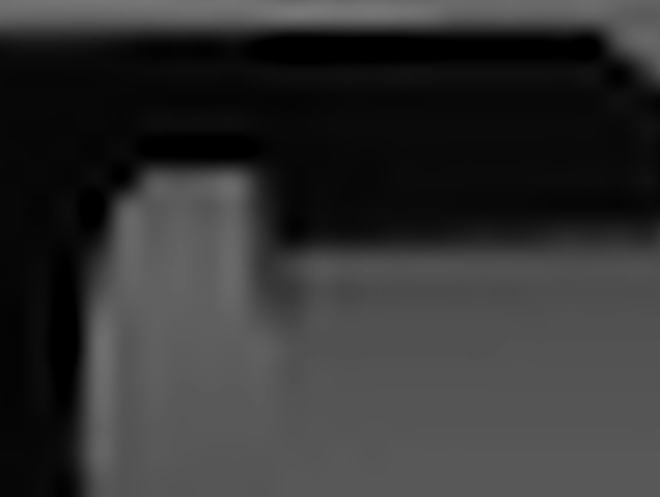}
\subcaption*{CMSR} 
\end{subfigure} \\
\begin{subfigure}{0.19\linewidth}
\includegraphics[width=1\linewidth]{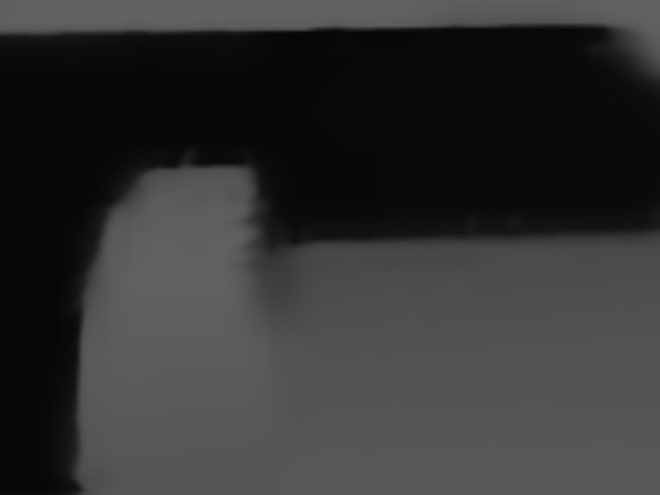}
\subcaption*{SSGNet*}
\end{subfigure}
\begin{subfigure}{0.19\linewidth}
\includegraphics[width=1\linewidth]{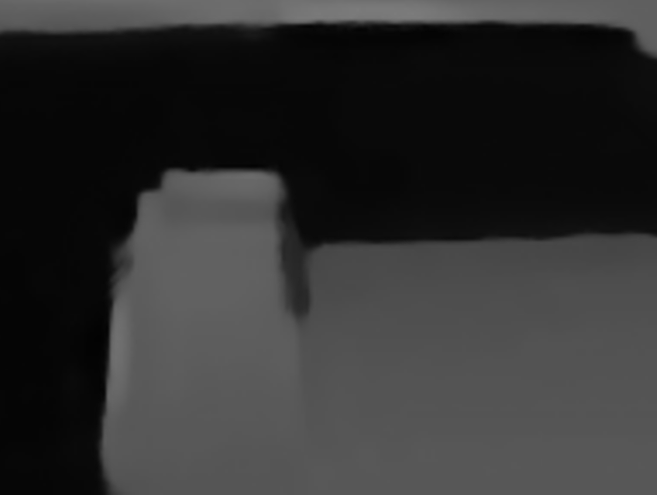}
\subcaption*{C2PD*}
\end{subfigure}
\begin{subfigure}{0.19\linewidth}
\includegraphics[width=1\linewidth]{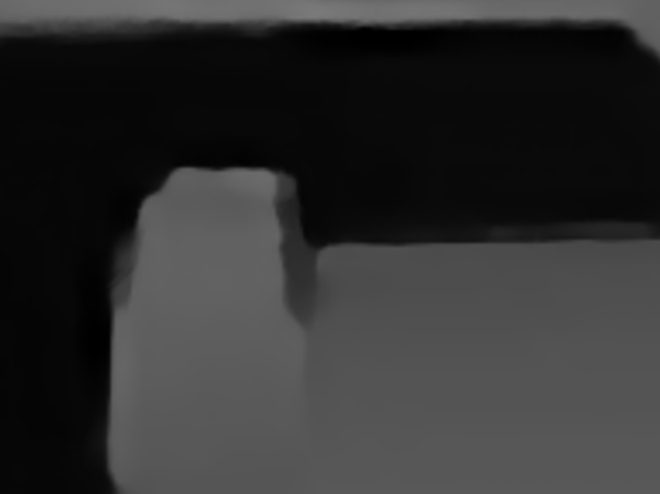}
\subcaption*{DORNet*}
\end{subfigure}
\begin{subfigure}{0.19\linewidth}
\includegraphics[width=1\linewidth]{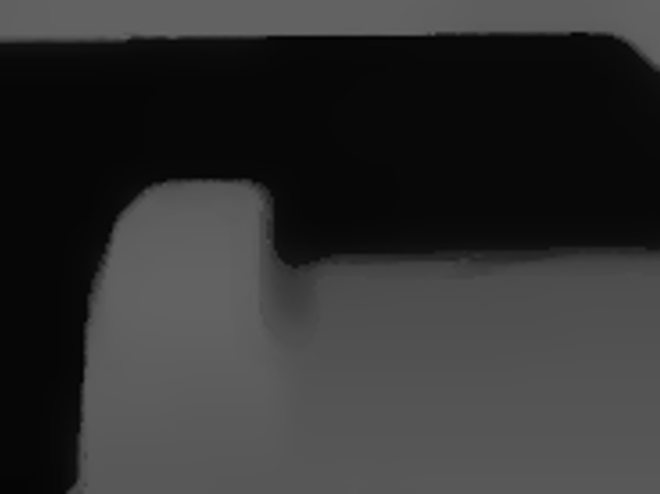}
\subcaption*{RobSelf-Re}
\end{subfigure}
\begin{subfigure}{0.19\linewidth}
\includegraphics[width=1\linewidth]{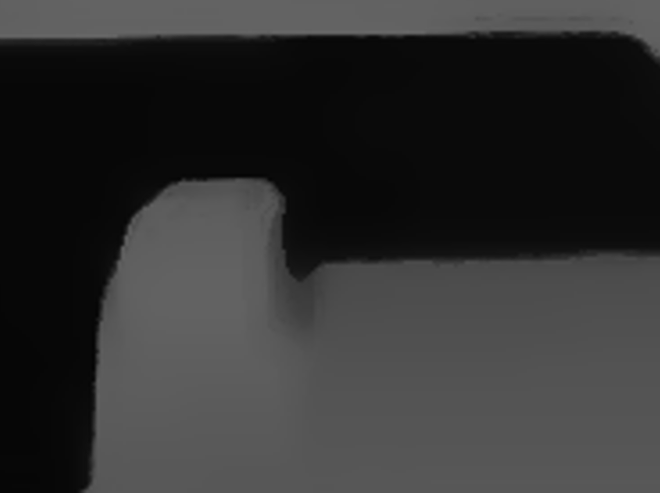}
\subcaption*{RobSelf-De}
\end{subfigure}
\caption{
Synthesized misaligned RGB-guided depth SR ($\times8$).
Image patches are shown due to space limit. 
*Pre-alignment applied.
}
\label{fig6:syn_rgb_depth}
\end{figure}

\subsection{Synthesized Misaligned RGB-Guided Depth SR} 
\label{subsec:syn_rgb_depth}

The estimator level $i$ and filter kernel sizes $\{m,n\}$ are set to $4$ and $\{7,5\}$ for $\times 4$ SR and $5$ and $\{13,7\}$ for $\times 8$ SR, with the filter threshold scaling factor $\eta=0.7$.
We compare with four supervised methods (DORNet~\cite{DORNet_2025_CVPR}, C2PD~\cite{CP2D_2025_AAAI}, SGNet~\cite{SGNet_2024_AAAI}, and DCTNet~\cite{DCTNet_2022_CVPR}) and four self-supervised methods (SSGNet~\cite{SSGNet_2023_AAAI}, MMSR~\cite{MMSR_ECCV_2022}, CMSR~\cite{CMSR_2021_CVPR}, and P2P~\cite{P2P_2019_ICCV}). 
The supervised methods are trained on the NYU v2 dataset~\cite{NYU_2012_ECCV}. 
Code of~\cite{MOMNet_2026} is unavailable.

\Cref{tab1_syn_rgb_depth} reports quantitative results. 
Our models achieve the best results for both SR factors,
despite requiring no training data, ground-truth supervision, or pre-alignment. 
\Cref{fig6:syn_rgb_depth} shows qualitative results.
CMSR~\cite{CMSR_2021_CVPR}, limited by its suboptimal alignment strategy that overlooks cross-modal dependencies, fails to effectively enhance the source. 
Two-stage methods (DORNet~\cite{DORNet_2025_CVPR}, C2PD~\cite{CP2D_2025_AAAI}, SSGNet~\cite{SSGNet_2023_AAAI}) perform better due to pre-alignment, but produce some non-faithful structures because their modules are not robust to residual misalignment. 
In comparison, our models deliver HR and high-fidelity results, 
approaching the ground truth. 
These improvements are attributed to our translator and filter, which jointly ensure effective guide alignment and faithful source enhancement.
\Cref{subsec:main_ablation} further analyzes the effect of our techniques.


\begin{table}[!t]
  \caption{
  Real-World misaligned RGB-guided depth SR.
  For methods lacking alignment strategies, results without and with pre-alignment are reported (wo/w).
  $^*$Params vary with SR factors ($\times2$/$\times4$) due to level $i$.
  The \textbf{best} and {\ul second best} are highlighted.
  } 
  \label{tab2_syn_rgb_depth}
  \centering
  \scalebox{0.61}{
  \begin{tabular}{l|cccccccc|cc}
    \toprule
     & P2P~\cite{P2P_2019_ICCV} & CMSR~\cite{CMSR_2021_CVPR} & DCTNet~\cite{DCTNet_2022_CVPR} & MMSR~\cite{MMSR_ECCV_2022} & SSGNet~\cite{SSGNet_2023_AAAI} & SGNet~\cite{SGNet_2024_AAAI} & DCNAS~\cite{DCNAS_2025_TPAMI} & DORNet~\cite{DORNet_2025_CVPR} & RobSelf-Re & RobSelf-De \\ \hline 
    Self-Supervised & $\checkmark$ & $\checkmark$ & \ding{55} & $\checkmark$ & $\checkmark$ & \ding{55} & \ding{55} & \ding{55} & $\checkmark$ & $\checkmark$ \\ 
    Pre-Alignment & $\checkmark$ & \ding{55} & $\checkmark$ & $\checkmark$ & $\checkmark$ & $\checkmark$ & $\checkmark$ & $\checkmark$ & \ding{55} & \ding{55} \\
    Params (M) & 0.2 & 1.18 & 0.48 & 0.25 & 0.31 & 9.22 & 0.60 & 3.05 & 0.65/0.76$^*$ & 0.70/0.81$^*$ \\ \hline
    $\times2$ (RMSE$\downarrow$) & 3.75/3.51 & 2.53 & 3.27/3.14 & 2.93/2.76 & 2.93/2.74 & 3.18/2.98 & 3.49/3.27 & 3.24/3.04 & {\ul 2.23} & \textbf{2.18} \\
    $\times4$ (DSS$\uparrow$) & 1.7/2.2 & 6.7 & 4.1/4.5 & 5.0/5.6 & 5.0/5.5 & 4.7/5.1 & 3.6/4.1 & 4.3/4.9 & {\ul 8.4} & \textbf{8.6} \\
    \bottomrule
  \end{tabular}
  } 
\end{table}

\begin{figure}[!t]
\centering
\begin{subfigure}{0.19\linewidth} 
\includegraphics[width=1\linewidth]{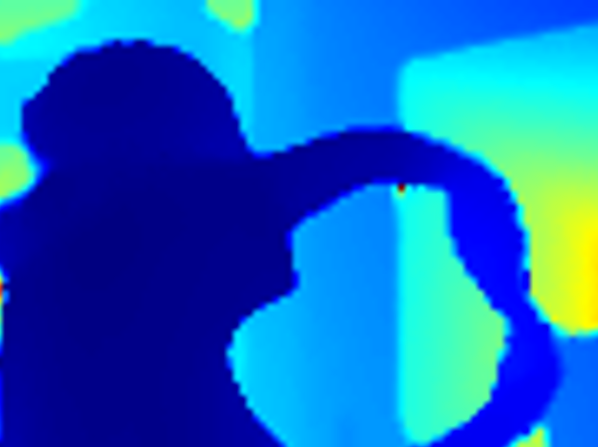}
\subcaption*{LR Source}
\end{subfigure}
\begin{subfigure}{0.19\linewidth}
\includegraphics[width=1\linewidth]{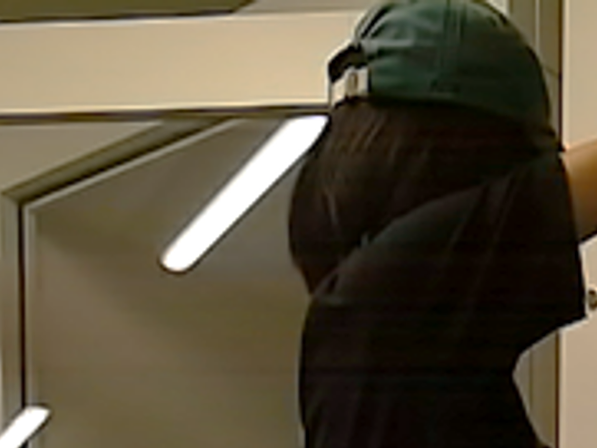}
\subcaption*{HR Guide}
\end{subfigure}
\begin{subfigure}{0.19\linewidth}
\includegraphics[width=1\linewidth]{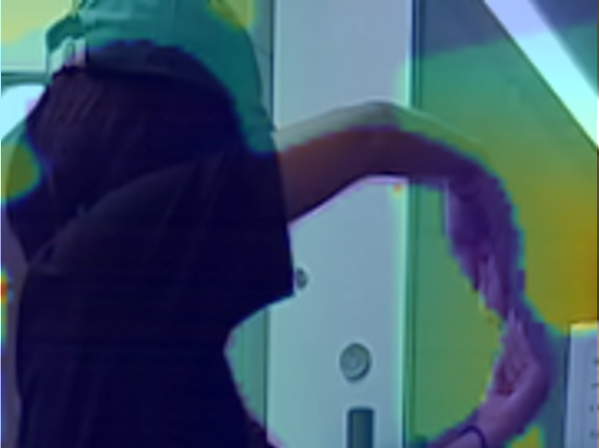}
\subcaption*{Pre-Aligned Guide}
\end{subfigure}
\begin{subfigure}{0.19\linewidth}
\includegraphics[width=1\linewidth]{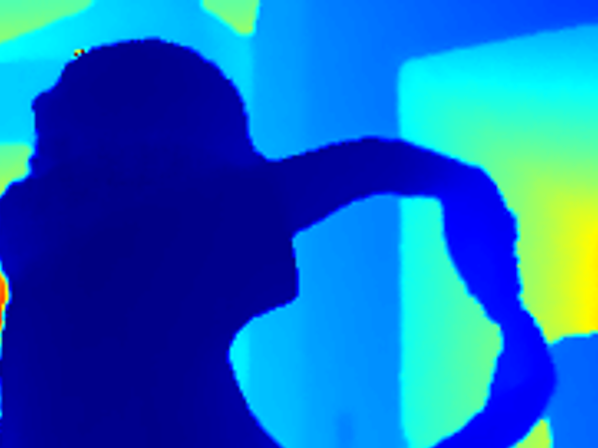}
\subcaption*{Ground Truth}
\end{subfigure}
\begin{subfigure}{0.19\linewidth}
\includegraphics[width=1\linewidth]{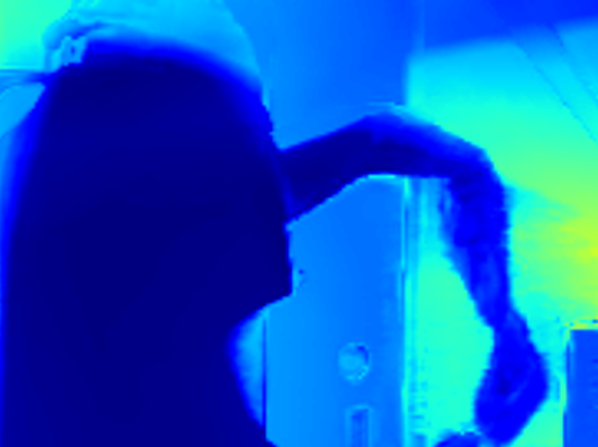}
\subcaption*{P2P*} 
\end{subfigure} \\
\begin{subfigure}{0.19\linewidth}
\includegraphics[width=1\linewidth]{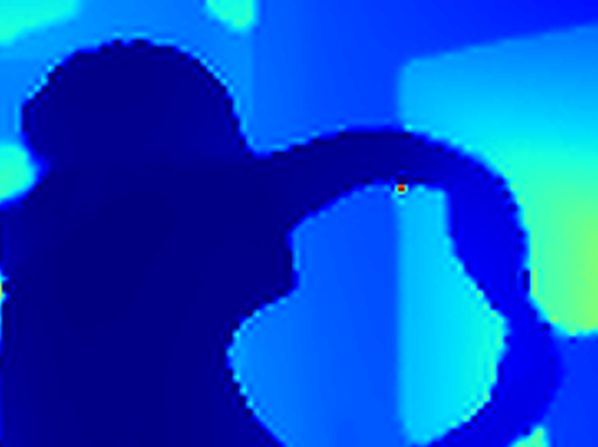}
\subcaption*{CMSR} 
\end{subfigure} 
\begin{subfigure}{0.19\linewidth}
\includegraphics[width=1\linewidth]{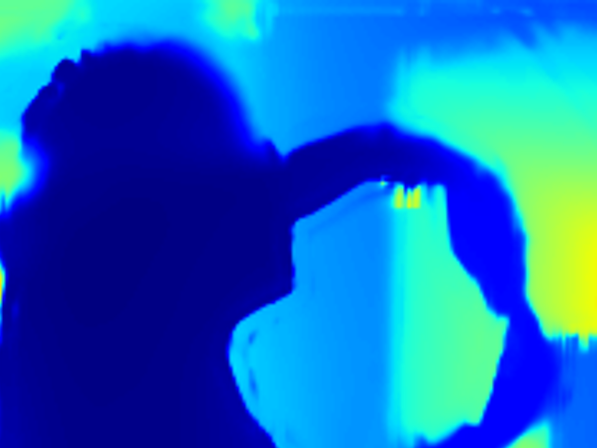}
\subcaption*{MMSR*} 
\end{subfigure}
\begin{subfigure}{0.19\linewidth}
\includegraphics[width=1\linewidth]{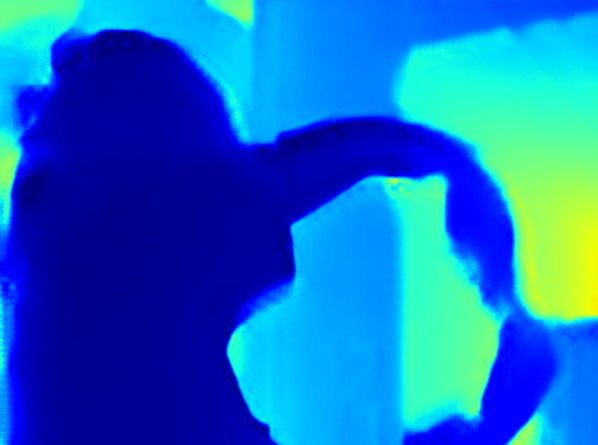}
\subcaption*{SGNet*} 
\end{subfigure}
\begin{subfigure}{0.19\linewidth}
\includegraphics[width=1\linewidth]{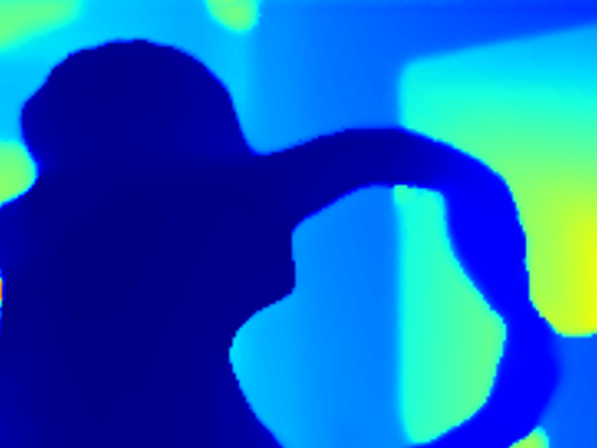}
\subcaption*{RobSelf-Re}
\end{subfigure}
\begin{subfigure}{0.19\linewidth}
\includegraphics[width=1\linewidth]{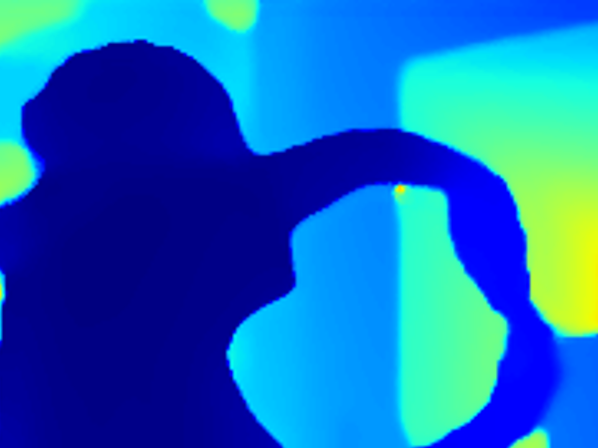}
\subcaption*{RobSelf-De} 
\end{subfigure} \\
\begin{subfigure}{0.19\linewidth} 
\includegraphics[width=1\linewidth]{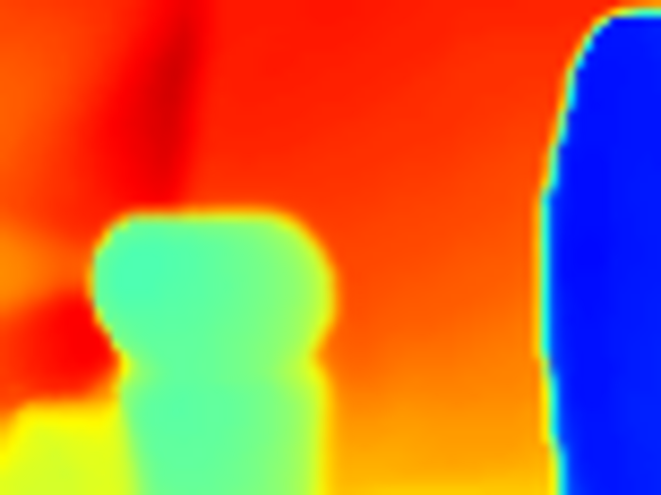}
\subcaption*{LR Source}
\end{subfigure}
\begin{subfigure}{0.19\linewidth}
\includegraphics[width=1\linewidth]{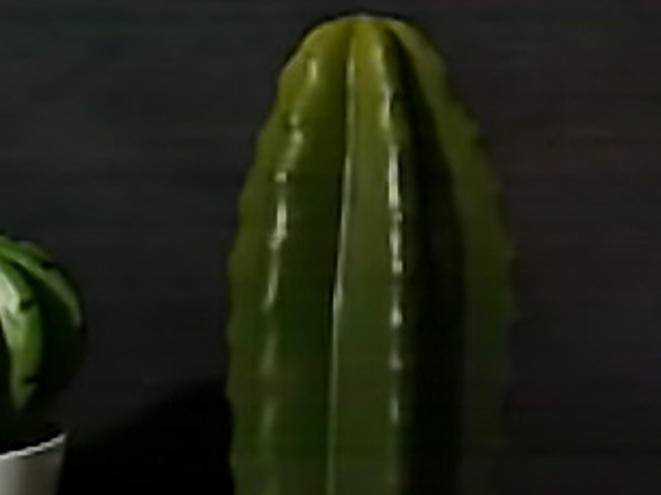}
\subcaption*{HR Guide}
\end{subfigure}
\begin{subfigure}{0.19\linewidth}
\includegraphics[width=1\linewidth]{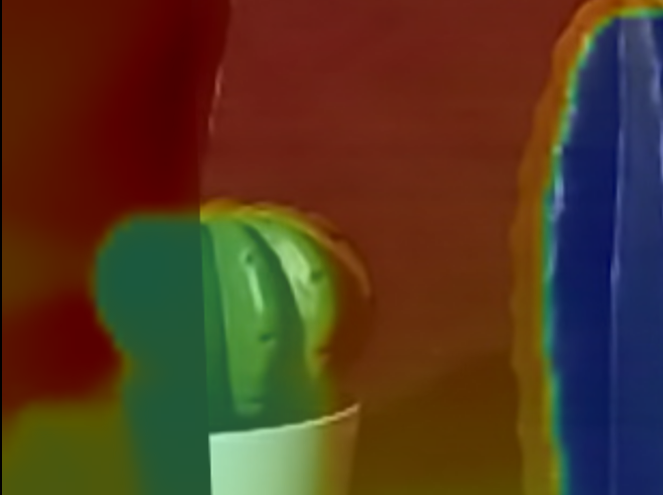}
\subcaption*{Pre-Aligned Guide}
\end{subfigure}
\begin{subfigure}{0.19\linewidth}
\includegraphics[width=1\linewidth]{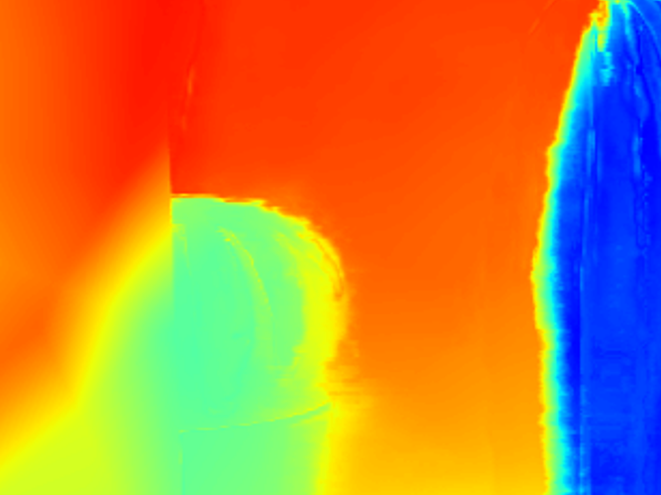}
\subcaption*{P2P*}
\end{subfigure}
\begin{subfigure}{0.19\linewidth}
\includegraphics[width=1\linewidth]{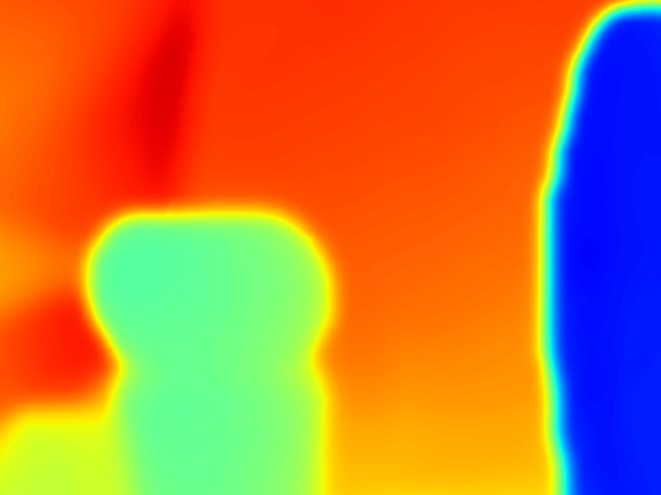}
\subcaption*{MMSR*} 
\end{subfigure} \\
\begin{subfigure}{0.19\linewidth}
\includegraphics[width=1\linewidth]{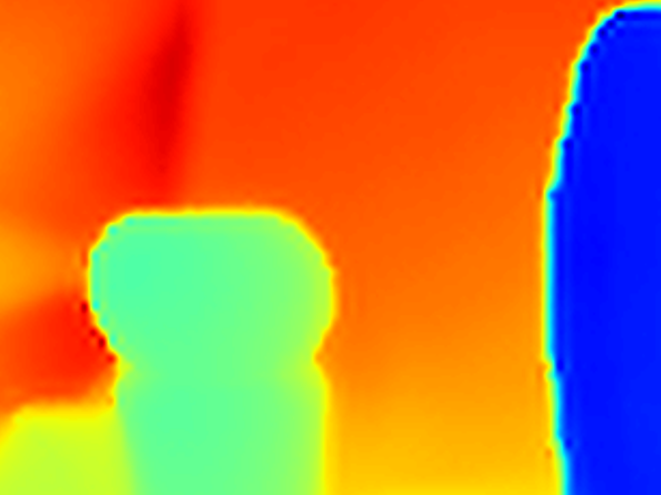}
\subcaption*{CMSR} 
\end{subfigure} 
\begin{subfigure}{0.19\linewidth}
\includegraphics[width=1\linewidth]{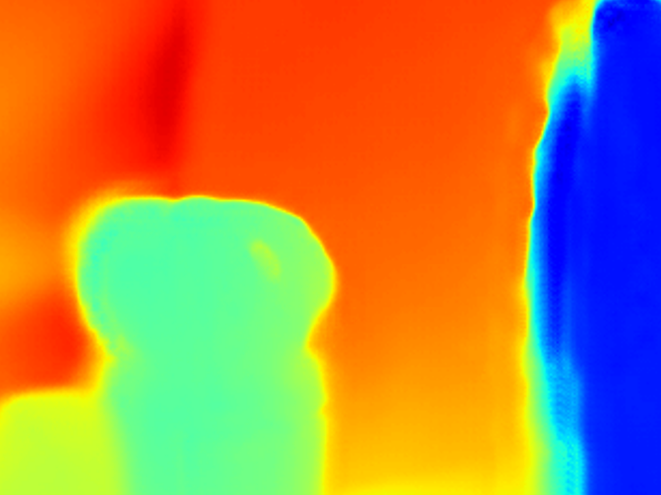}
\subcaption*{SGNet*} 
\end{subfigure}
\begin{subfigure}{0.19\linewidth}
\includegraphics[width=1\linewidth]{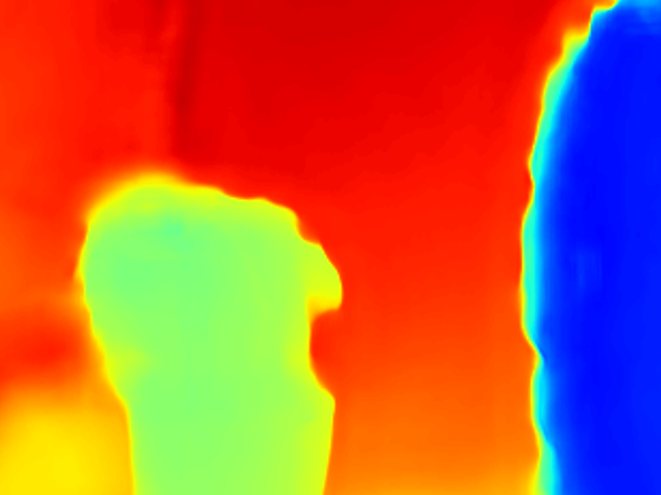}
\subcaption*{DORNet*}
\end{subfigure}
\begin{subfigure}{0.19\linewidth}
\includegraphics[width=1\linewidth]{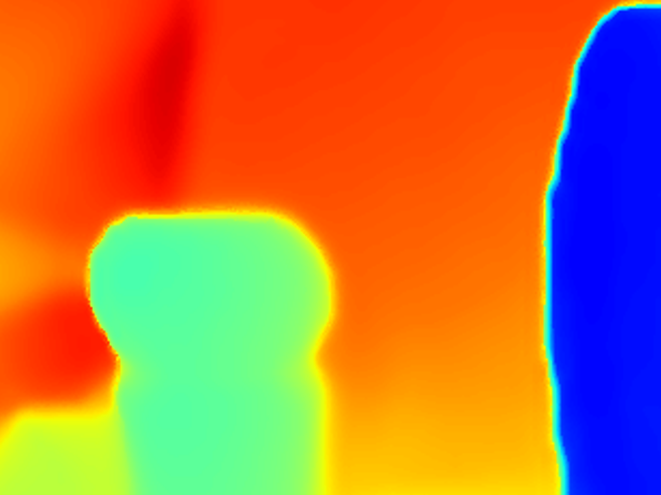}
\subcaption*{RobSelf-Re}
\end{subfigure}
\begin{subfigure}{0.19\linewidth}
\includegraphics[width=1\linewidth]{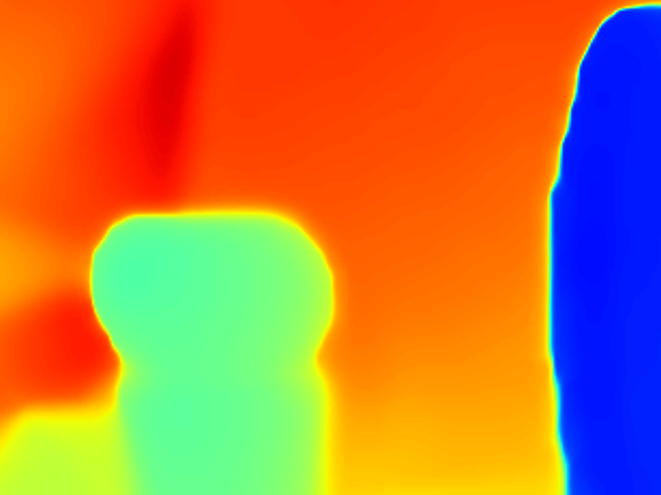}
\subcaption*{RobSelf-De}
\end{subfigure} 
\caption{
Real-World misaligned RGB-guided depth SR (upper: $\times2$; lower: $\times4$).
Image patches are shown due to space limit.
*Pre-alignment applied.
}
\label{fig7:real_rgb_depth} 
\end{figure}

\subsection{Real-World Misaligned RGB-Guided Depth SR} 
\label{subsec:real_rgb_depth}

The estimator level $i$ is set to $3$ for $\times 2$ SR and $4$ for $\times 4$ SR.
The filter threshold scaling factor $\eta$ and kernel sizes $\{m,n\}$ are fixed to 0.7 and $\{7,5\}$, respectively.
We compare with four supervised methods (DORNet~\cite{DORNet_2025_CVPR}, DCNAS~\cite{DCNAS_2025_TPAMI}, SGNet~\cite{SGNet_2024_AAAI}, and DCTNet~\cite{DCTNet_2022_CVPR}) and four self-supervised methods (SSGNet~\cite{SSGNet_2023_AAAI}, MMSR~\cite{MMSR_ECCV_2022}, CMSR~\cite{CMSR_2021_CVPR}, and P2P~\cite{P2P_2019_ICCV}). 
The supervised methods are trained on the RGB-D-D dataset~\cite{FDSR_2021_CVPR}.  

\Cref{tab2_syn_rgb_depth} shows quantitative results. 
On real-world misaligned RGB-depth data, our models outperform other methods by a large margin across SR factors. 
\Cref{fig7:real_rgb_depth} shows qualitative results.
Two-stage methods suffer from poor boundaries (DORNet~\cite{DORNet_2025_CVPR}), ghosting artifacts (SGNet~\cite{SGNet_2024_AAAI}), or spurious textures (P2P~\cite{P2P_2019_ICCV}), yielding less enhancement here than on synthesized data. 
This is because the pre-alignment struggles to generalize to such real-world data, where the misalignment is more complex due to various cross-sensor discrepancies and viewpoint variation. 
By contrast, our models achieve robustness and generalizability on real-world data with complex, large misalignments,
owing to the joint weakly-supervised, misalignment-aware translation formulation~(\cref{subsec:interesting}).   
See the supplementary material for analysis of pre-alignment. 


\begin{table}[!t]
  \caption{
  Real-World misaligned RGB-guided NIR SR.
  For methods lacking alignment strategies, results without and with pre-alignment are reported (wo/w).
  $^*$Params vary with SR factors ($\times2$/$\times4$) due to level $i$.
  The \textbf{best} and {\ul second best} are highlighted.
  } 
  \label{tab3_real_rgb_nir}
  \centering
  \scalebox{0.59}{  
  \begin{tabular}{l|cccccccc|cc}
    \toprule
     & P2P~\cite{P2P_2019_ICCV} & CMSR~\cite{CMSR_2021_CVPR} 
     & MMSR~\cite{MMSR_ECCV_2022} & SSGNet~\cite{SSGNet_2023_AAAI} 
     & Fusion-ArF~\cite{Fusion_2025_TIP} & Fusion-CAP~\cite{MulFS-CAP_2025_PAMI} & PAN~\cite{PAN_2025_ICCV} & HSI SR~\cite{UnalignedHSI_2025_IJCV} 
     & RobSelf-Re & RobSelf-De \\ \hline     
    Self-Supervised & $\checkmark$ & $\checkmark$ & $\checkmark$ & $\checkmark$ 
    & \ding{55} & \ding{55} & \ding{55} & \ding{55} 
    & $\checkmark$ & $\checkmark$ \\ 
    Pre-Alignment & $\checkmark$ & \ding{55}  & $\checkmark$ & $\checkmark$ 
    & \ding{55} & \ding{55} & \ding{55} & \ding{55} 
    & \ding{55} & \ding{55} \\
    Params (M) & 0.2 & 1.18 & 0.25 & 0.31 
    & 47.39 & 0.38 & 7.17 & 10.97 
    & 0.91/1.02$^*$ & 0.96/1.07$^*$ \\ \hline 
    $\times2$ (RMSE$\downarrow$)  & 18.72/18.91 & 3.42 & 8.94/8.91 & 8.96/9.04
    & 6.58 & 4.40 & 5.19 & 4.52 
    & \textbf{3.09} & {\ul 3.12} \\ 
    $\times4$ (NIQE$\downarrow$)  & 6.32/\textbf{5.32} & 8.05 & 10.95/10.50 & 10.30/10.23 
    & 9.54 & 8.95 & 9.24 & 9.01 
    & 7.63 & {\ul 7.60} \\
    \bottomrule
  \end{tabular}
  } 
\end{table}

\begin{figure}[t]
\centering
\begin{subfigure}{0.19\linewidth} 
\includegraphics[width=1\linewidth]{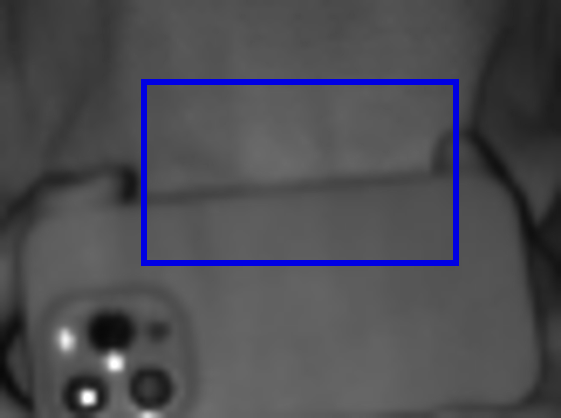}
\subcaption*{LR Source}
\end{subfigure}
\begin{subfigure}{0.19\linewidth}
\includegraphics[width=1\linewidth]{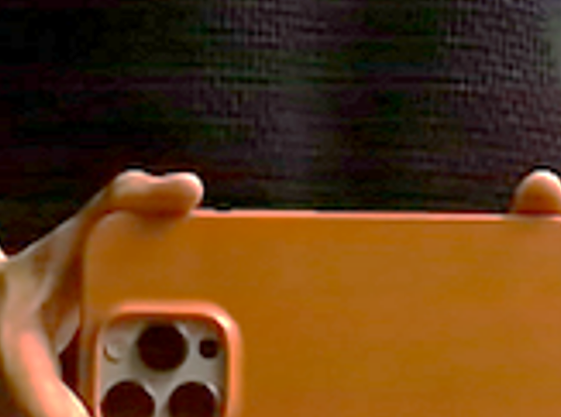}
\subcaption*{HR Guide}
\end{subfigure}
\begin{subfigure}{0.19\linewidth}
\includegraphics[width=1\linewidth]{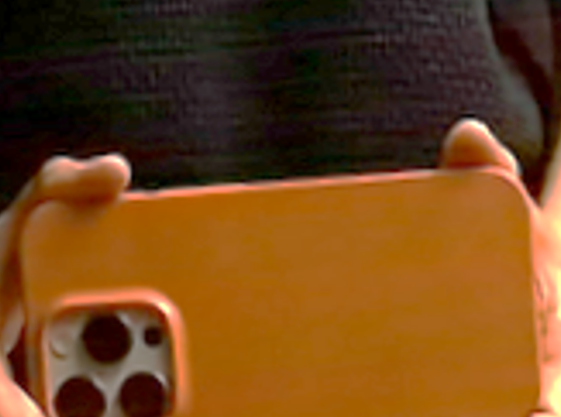}
\subcaption*{Pre-Aligned Guide}
\end{subfigure}
\begin{subfigure}{0.19\linewidth}
\includegraphics[width=1\linewidth]{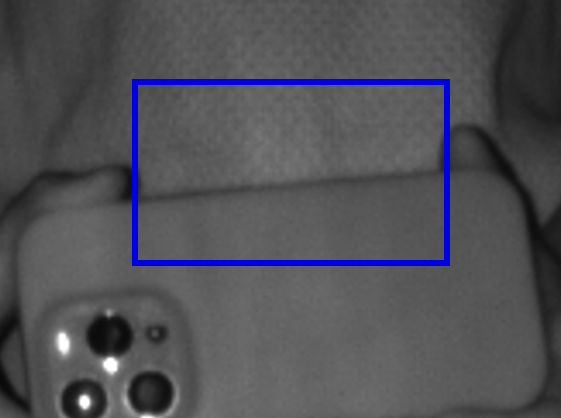}
\subcaption*{Ground Truth}
\end{subfigure}
\begin{subfigure}{0.19\linewidth}
\includegraphics[width=1\linewidth]{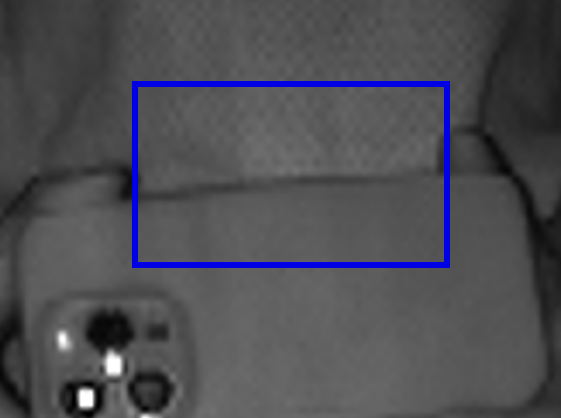}
\subcaption*{CMSR}
\end{subfigure}
\begin{subfigure}{0.19\linewidth}
\includegraphics[width=1\linewidth]{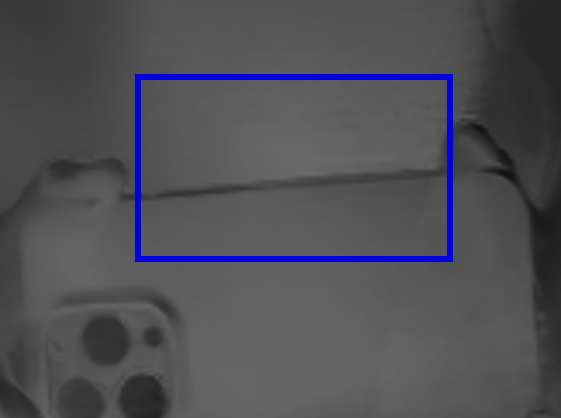}
\subcaption*{P2P*}
\end{subfigure}
\begin{subfigure}{0.19\linewidth}
\includegraphics[width=1\linewidth]{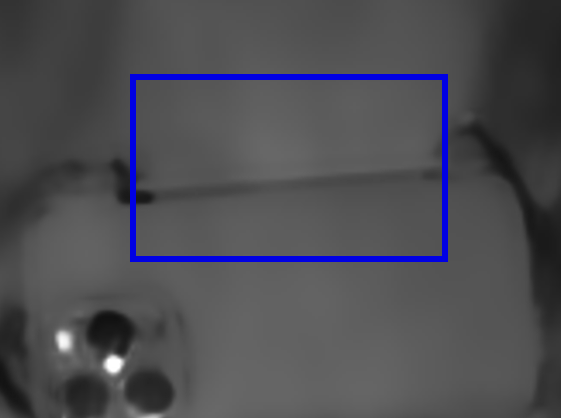}
\subcaption*{SSGNet*}
\end{subfigure}
\begin{subfigure}{0.19\linewidth}
\includegraphics[width=1\linewidth]{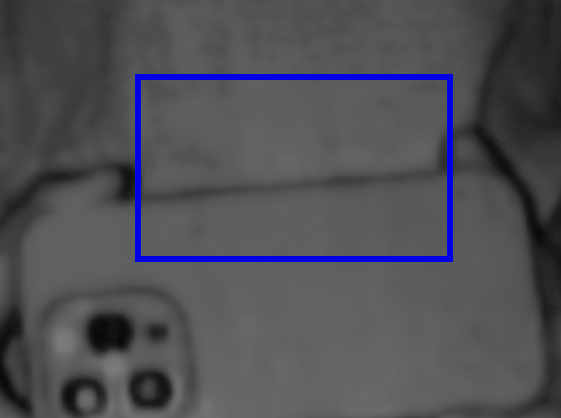}
\subcaption*{Fusion-CAP}
\end{subfigure}
\begin{subfigure}{0.19\linewidth} 
\includegraphics[width=1\linewidth]{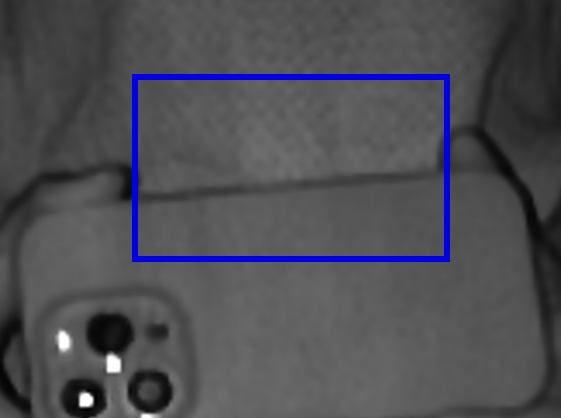}
\subcaption*{RobSelf-Re}
\end{subfigure}
\begin{subfigure}{0.19\linewidth}
\includegraphics[width=1\linewidth]{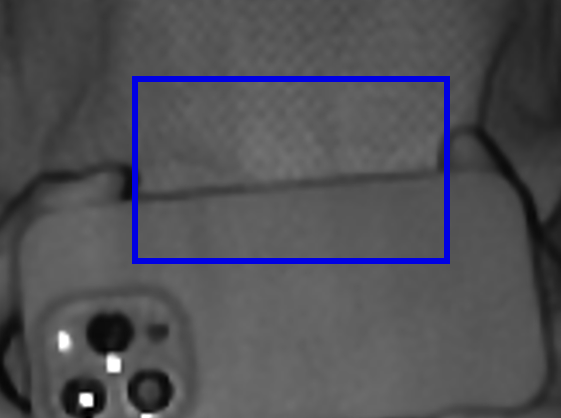}
\subcaption*{RobSelf-De}
\end{subfigure}
\caption{
Real-world misaligned RGB-guided NIR SR ($\times 2$). 
Image patches are shown due to space limit. 
*Pre-alignment applied.
}
\label{fig8_real_rgb_nir} 
\end{figure}

\subsection{Real-World Misaligned RGB-Guided NIR SR}  
\label{subsec:real_rgb_nir}

The estimator level $i$ is set to 4 for $\times 2$ SR and 5 for $\times 4$ SR. 
The filter kernel sizes $\{m,n\}$ are both set to 3 to better capture the NIR details, with no threshold applied.
We compare with prior self-supervised methods (SSGNet~\cite{SSGNet_2023_AAAI}, MMSR~\cite{MMSR_ECCV_2022}, CMSR~\cite{CMSR_2021_CVPR}, P2P~\cite{P2P_2019_ICCV}) 
and misalignment-aware, supervised methods for guided HSI SR~\cite{UnalignedHSI_2025_IJCV}, pan-sharpening~\cite{PAN_2025_ICCV}, and fusion~\cite{MulFS-CAP_2025_PAMI,Fusion_2025_TIP}.
The supervised methods are trained on EPFL RGB-NIR~\cite{EPFL_2011_CVPR} with synthesized misalignment. 

\Cref{tab3_real_rgb_nir} and \Cref{fig8_real_rgb_nir} show quantitative and qualitative results, respectively.
This task is more challenging due to object motion and the rich NIR details. 
Other methods degrade numerically and suffer from 
blur (Fusion-CAP~\cite{MulFS-CAP_2025_PAMI}, SSGNet~\cite{SSGNet_2023_AAAI}, P2P~\cite{P2P_2019_ICCV}) or edge distortion (CMSR~\cite{CMSR_2021_CVPR}). 
In contrast, our models achieve more effective enhancement and recover finer details, thanks to our translator and filter which together enable robust handling of diverse real-world misalignments and effective reconstruction of fine NIR details. 
See the supplementary material for more qualitative results. 


\begin{figure}[t]
\centering
\begin{subfigure}{0.14\linewidth} 
\includegraphics[width=1\linewidth]{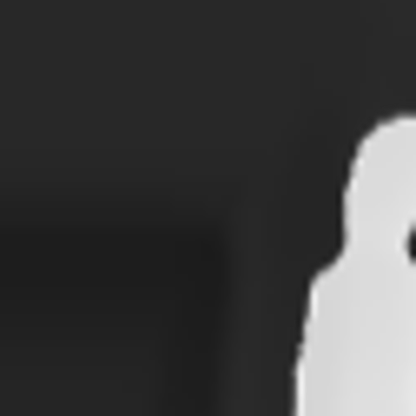}
\subcaption*{LR Source}
\end{subfigure}
\begin{subfigure}{0.14\linewidth}
\includegraphics[width=1\linewidth]{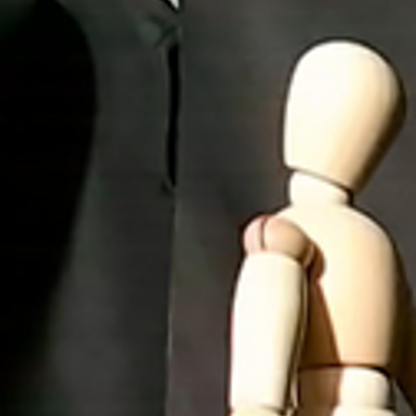}
\subcaption*{HR Guide} 
\end{subfigure}
\begin{subfigure}{0.14\linewidth}
\includegraphics[width=1\linewidth]{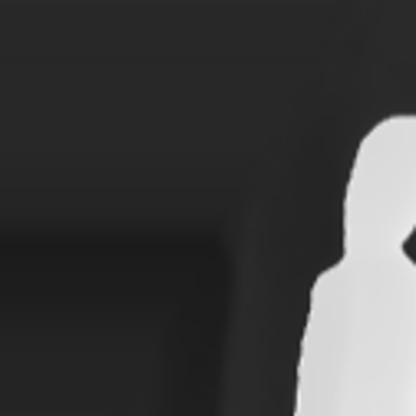}
\subcaption*{Ground Truth}
\end{subfigure} \\
\begin{subfigure}{0.14\linewidth}
\includegraphics[width=1\linewidth]{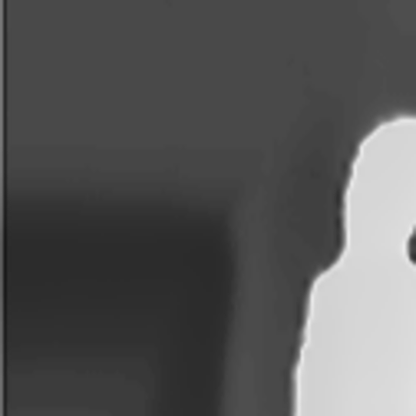}
\subcaption*{w/o T and F}
\end{subfigure}
\begin{subfigure}{0.14\linewidth}
\includegraphics[width=1\linewidth]{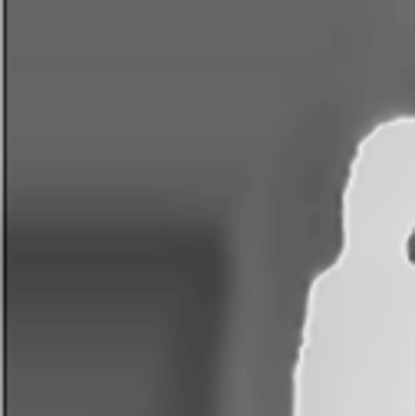}
\subcaption*{w/ only T} 
\end{subfigure}
\begin{subfigure}{0.14\linewidth}
\includegraphics[width=1\linewidth]{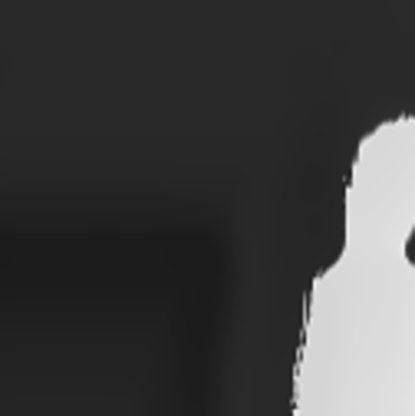}
\subcaption*{w/ only F}
\end{subfigure}
\begin{subfigure}{0.14\linewidth}
\includegraphics[width=1\linewidth]{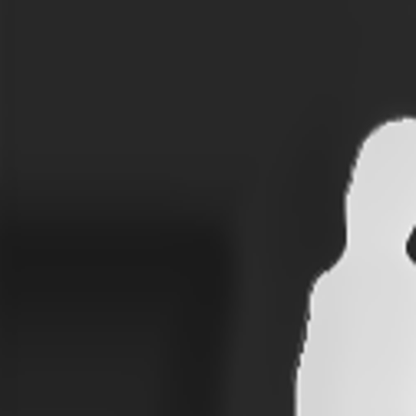}
\subcaption*{w/ T and F} 
\end{subfigure} \\
\begin{subfigure}{0.13\linewidth}
\includegraphics[width=1\linewidth]{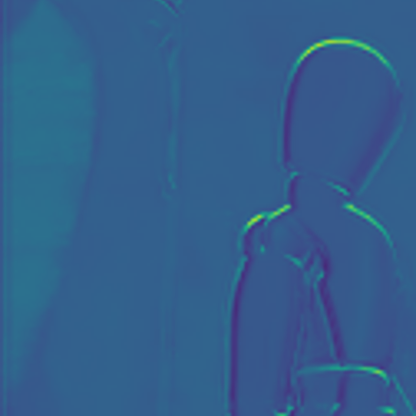}
\subcaption*{$\mathbf{F}_{guide}$} 
\end{subfigure} 
\begin{subfigure}{0.13\linewidth}
\includegraphics[width=1\linewidth]{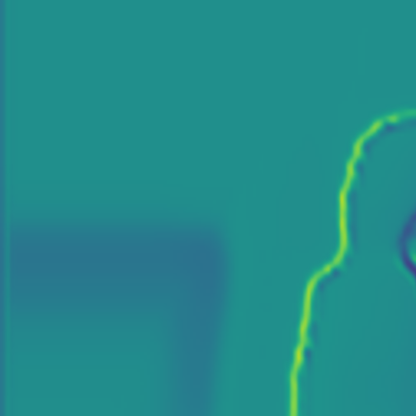}
\subcaption*{$\mathbf{F}_{source}$} 
\end{subfigure}
\begin{subfigure}{0.128\linewidth}
\includegraphics[width=1\linewidth]{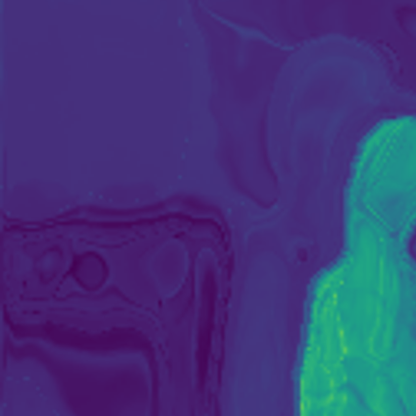}
\subcaption*{\tiny $\mathbf{F}^{\rm Aligned}_{guide}$} 
\end{subfigure}
\begin{subfigure}{0.13\linewidth}
\includegraphics[width=1\linewidth]{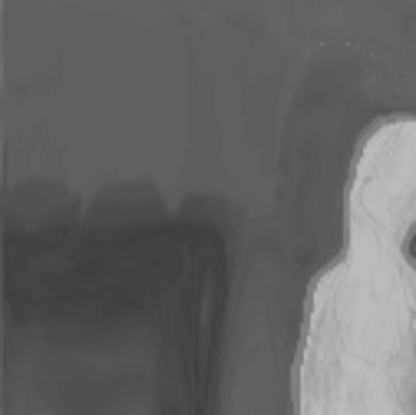}
\subcaption*{$\mathbf{I}^{\rm Trans}_{pred}$} 
\end{subfigure}
\begin{subfigure}{0.13\linewidth}
\includegraphics[width=1\linewidth]{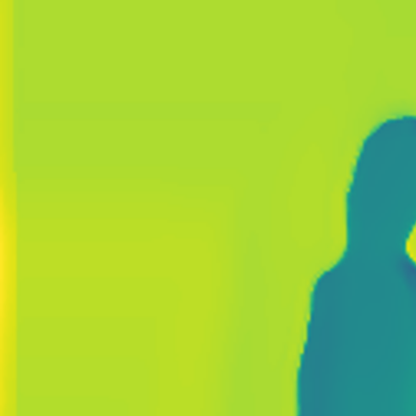}
\subcaption*{$\mathbf{F}^{\rm Enhanced}_{source}$} 
\end{subfigure} 
\caption{ 
Effectiveness of our translator and filter.
}
\label{fig9_ablation}
\end{figure}

\begin{table}[!t]
  \caption{
  Effect of our translator and filter. 
  For first two settings, $\{\mathbf{F}_{source}, \mathbf{F}_{guide}\}$ or $\{\mathbf{F}_{source}, \mathbf{F}^{\rm Aligned}_{guide}\}$ are fused using a $1\times1$ conv.  
  No feature fusion in last two settings. 
  } 
  \label{tab4_ablation}
  \centering
  \scalebox{0.7}{ 
  \begin{tabular}{l|cccc}
    \toprule
    Translator & \ding{55} & $\checkmark$ & \ding{55} & $\checkmark$ \\
    Filter & \ding{55} & \ding{55} & $\checkmark$ & $\checkmark$ \\ \hline
    RMSE$\downarrow$ & 2.55 & 2.37 & 2.31 & \textbf{2.18} \\
    \bottomrule
  \end{tabular}
  } 
\end{table}

\subsection{Ablation Study}
\label{subsec:main_ablation}

We analyze the overall effect of our translator and filter using RobSelf-De on task \uppercase\expandafter{\romannumeral2} ($\times2$ SR).
As shown in \cref{tab4_ablation} and \cref{fig9_ablation} (upper), 
when both are removed, the result remains unenhanced since neither the misaligned guide nor self-enhancement is effectively exploited.
When only the translator is adopted, the result obtains weak enhancement due to the simple fusion of aligned guidance. 
When only the filter is adopted, the result obtains naive self-enhancement under misaligned guidance and introduces false structures. 
When the translator is further used alongside the filter 
to provide aligned guidance for reference-based self-enhancement, the result exhibits both high resolution and high fidelity. 
The bare baseline can outperform some previous methods because their modules overly fuse the misaligned guidance, which degrades source structures. This also motivates our design of reference-based self-enhancement. 

\Cref{fig9_ablation} (lower) visualizes the features from the full model.
Our translator, by warping $\mathbf{F}_{guide}$ and driving it to mimic source modality via weakly-supervised translation ($\mathbf{I}^{\rm Trans}_{pred}$), 
effectively derives $\mathbf{F}^{\rm Aligned}_{guide}$ aligned with $\mathbf{F}_{source}$. 
Since $\mathbf{F}^{\rm Aligned}_{guide}$ remains in the HR space and is supervised only by LR source, it preserves both essential source-corresponding structures and redundant modality-specific content. The latter can neither be aligned due to modality discrepancies nor removed during translation under weak supervision. 
Therefore, our filter uses $\mathbf{F}^{\rm Aligned}_{guide}$ only as reference to learn content-aware kernels for discriminative self-enhancement, which refines $\mathbf{F}_{source}$ into $\mathbf{F}^{\rm Enhanced}_{source}$, recovering fine details and avoiding redundancy effects.
Together, our techniques provide an effective inductive bias for robust self-supervised cross-modal SR under misalignment. 

See the supplementary material for more analyses on the translator and filter, as well as limitations. 


\begin{figure}[!t]
\centering
\begin{subfigure}{0.118\linewidth}
\subcaption*{$\mathbf{F}_{guide}$} 
\includegraphics[width=1\linewidth]{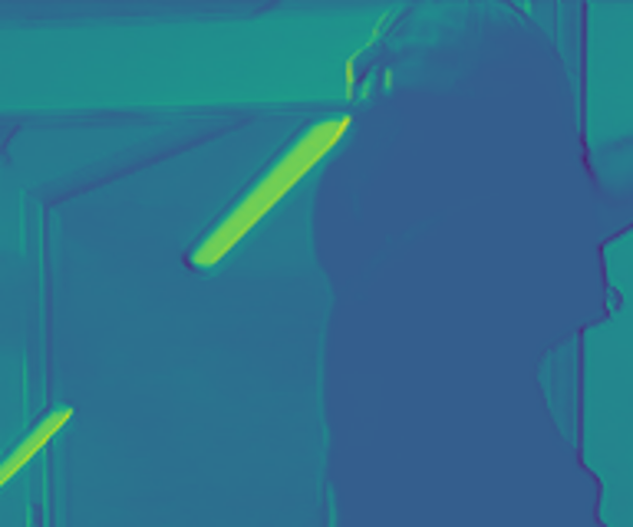}
\end{subfigure} 
\begin{subfigure}{0.118\linewidth} 
\subcaption*{$\mathbf{F}_{source}$} 
\includegraphics[width=1\linewidth]{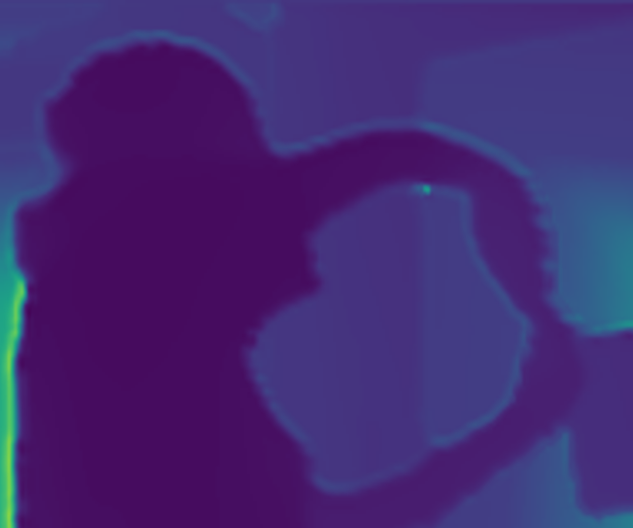}
\end{subfigure} 
\begin{subfigure}{0.118\linewidth}
\subcaption*{\tiny Modules~\cite{MURF_2023_PAMI}} 
\includegraphics[width=1\linewidth]{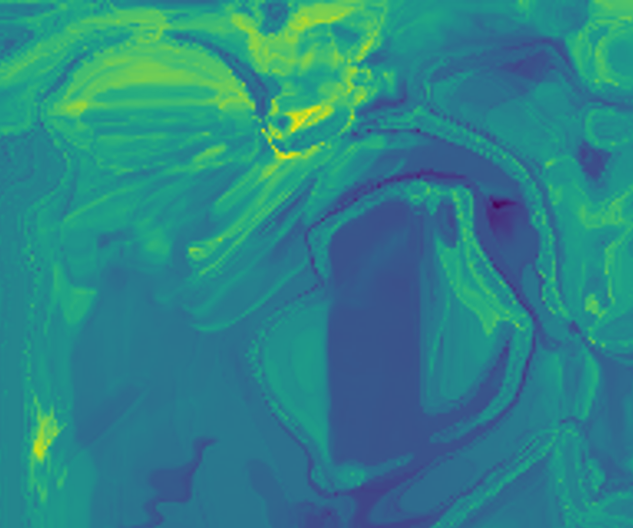}
\end{subfigure} 
\begin{subfigure}{0.118\linewidth}
\subcaption*{\tiny Translator}
\includegraphics[width=1\linewidth]{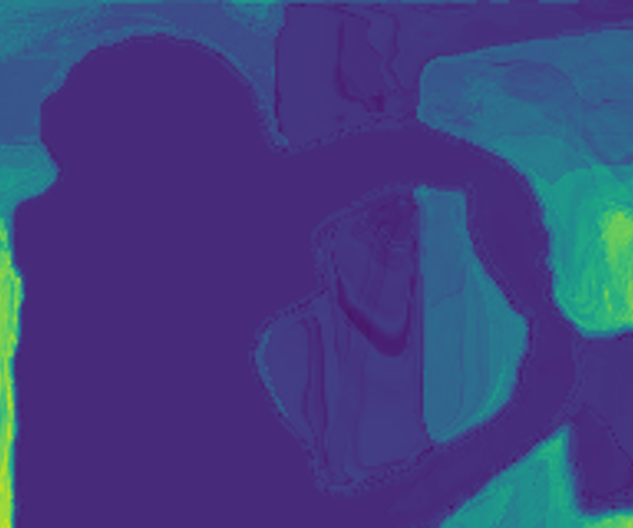}
\end{subfigure} 
\begin{subfigure}{0.118\linewidth}
\subcaption*{$\mathbf{F}_{guide}$} 
\includegraphics[width=1\linewidth]{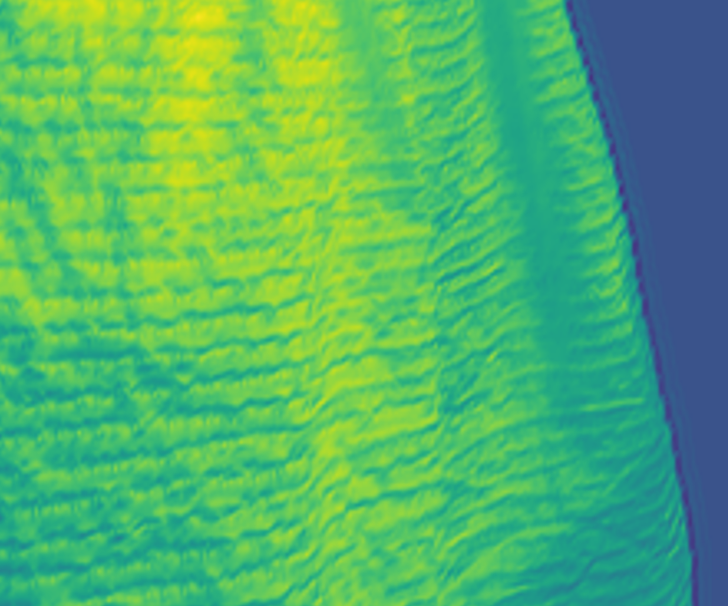}
\end{subfigure} 
\begin{subfigure}{0.118\linewidth} 
\subcaption*{$\mathbf{F}_{source}$} 
\includegraphics[width=1\linewidth]{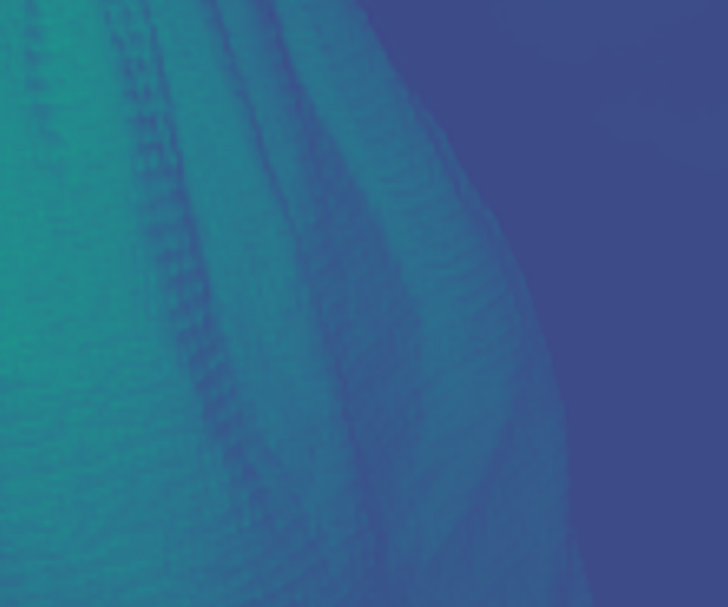}
\end{subfigure} 
\begin{subfigure}{0.118\linewidth}
\subcaption*{\tiny Modules~\cite{MURF_2023_PAMI}} 
\includegraphics[width=1\linewidth]{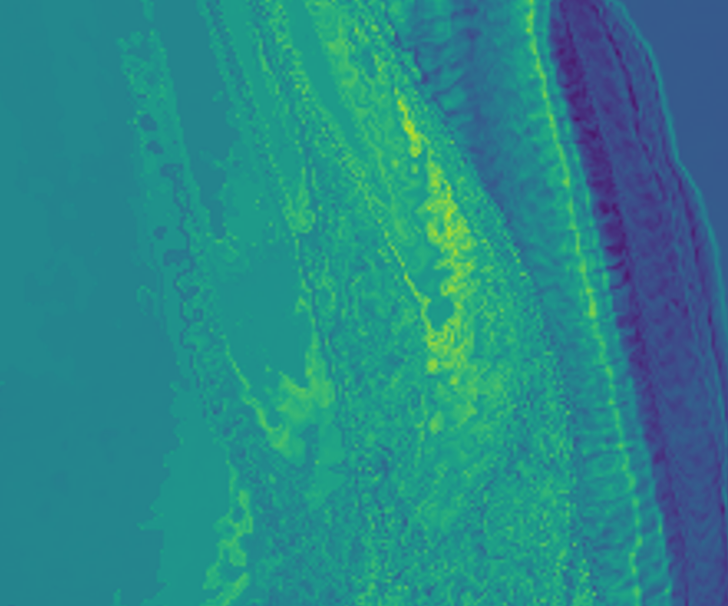}
\end{subfigure} 
\begin{subfigure}{0.118\linewidth}
\subcaption*{\tiny Translator}
\includegraphics[width=1\linewidth]{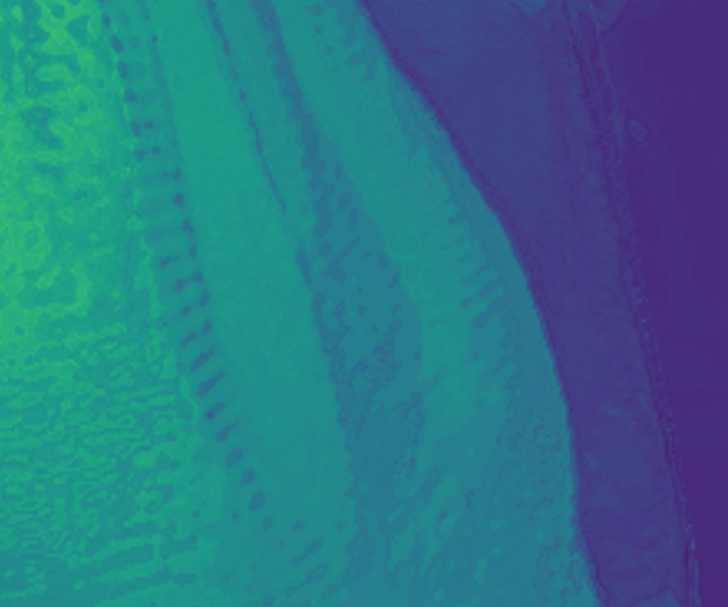}
\end{subfigure} 
\caption{
Guide alignment performance: our translator vs. modules in~\cite{MURF_2023_PAMI}.
Visualizations for real-world misaligned RGB-guided depth (left) and NIR (right) SR.
}
\label{supp_fig13_ga} 
\end{figure}

\begin{figure}[!t]
\centering
\begin{minipage}{0.26\linewidth} 
\includegraphics[width=1\linewidth]{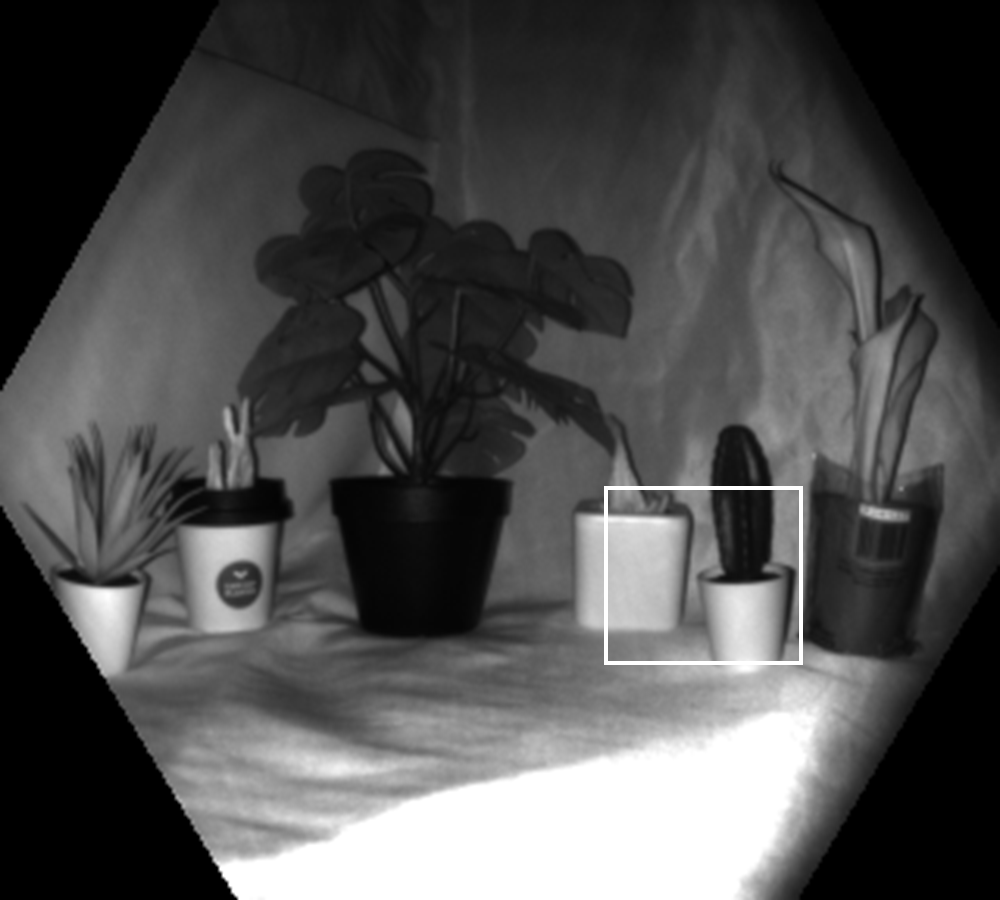}
\end{minipage}
\begin{minipage}{0.26\linewidth} 
\includegraphics[width=1\linewidth]{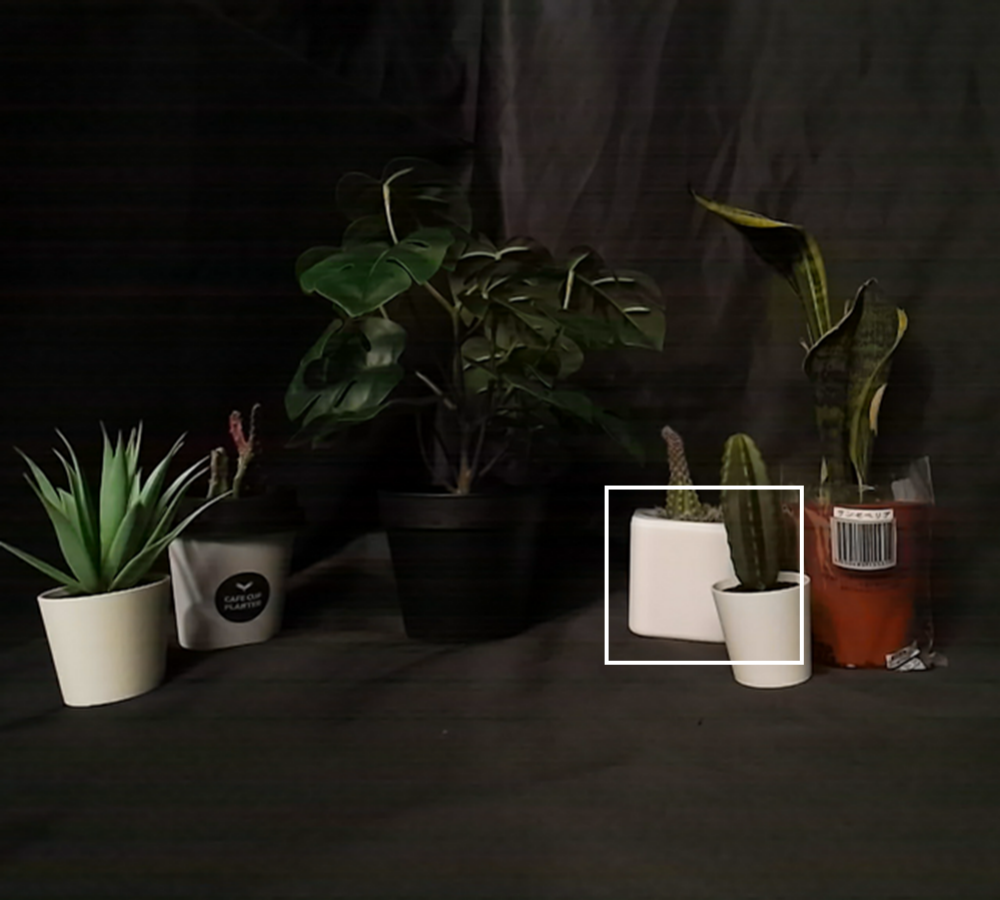}
\end{minipage}
\begin{minipage}{0.1\linewidth} 
\subcaption*{\tiny HR Guide} 
\includegraphics[width=1\linewidth]{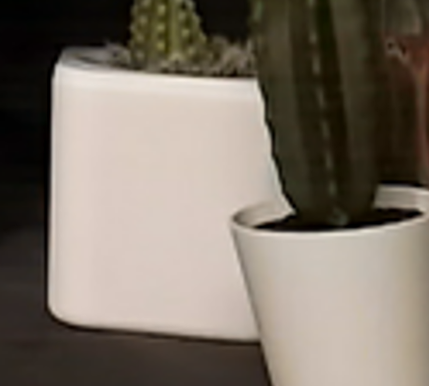}
\subcaption*{\tiny LR Source} 
\includegraphics[width=1\linewidth]{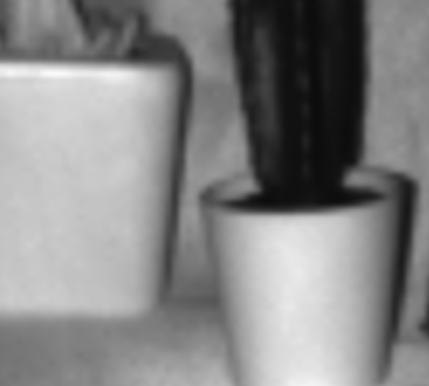}
\end{minipage}
\begin{minipage}{0.1\linewidth} 
\subcaption*{\tiny $\mathbf{F}_{guide}$} 
\includegraphics[width=1\linewidth]{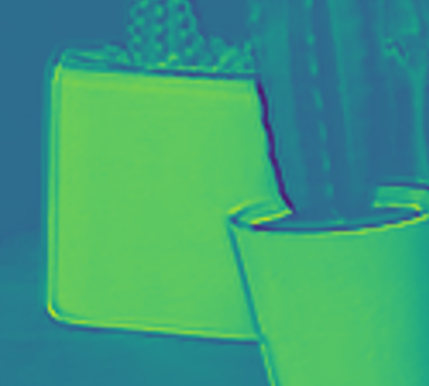}
\subcaption*{\tiny $\mathbf{F}_{source}$} 
\includegraphics[width=1\linewidth]{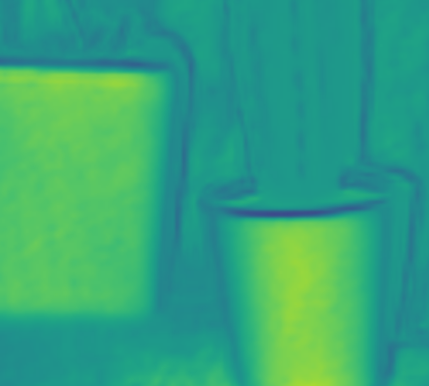}
\end{minipage}
\begin{minipage}{0.1\linewidth}
\subcaption*{\tiny $\mathbf{F}^{\rm Aligned}_{guide}$}
\includegraphics[width=1\linewidth]{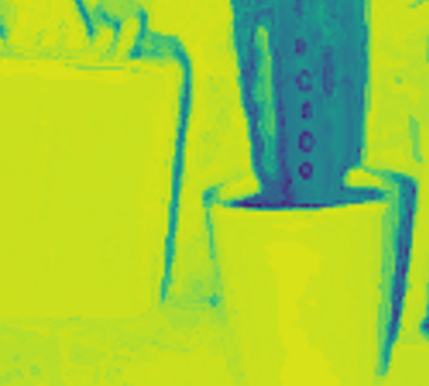}
\subcaption*{\tiny $\mathbf{F}^{\rm Enhanced}_{source}$} 
\includegraphics[width=1\linewidth]{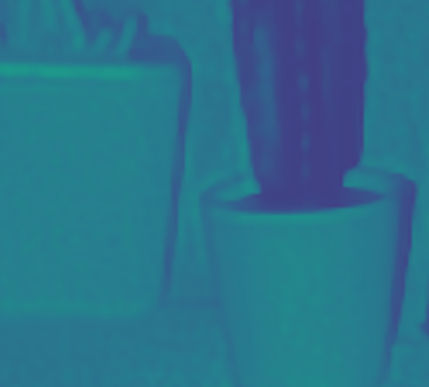}
\end{minipage}
\begin{minipage}{0.1\linewidth} 
\subcaption*{\tiny $\mathbf{I}^{\rm Trans}_{pred}$} 
\includegraphics[width=1\linewidth]{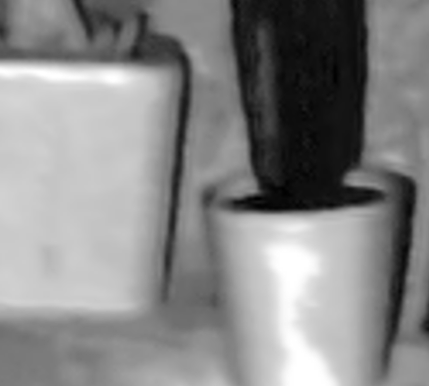}
\subcaption*{\tiny $\mathbf{I}^{\rm SR}_{pred}$} 
\includegraphics[width=1\linewidth]{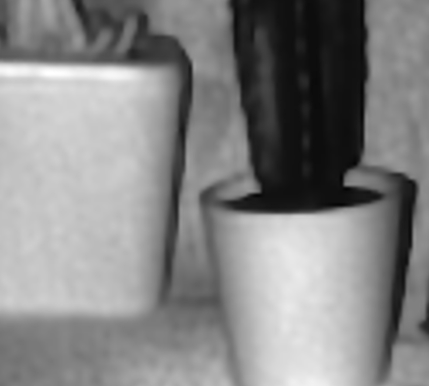}
\end{minipage}
\caption{
Our translator can ``synthesize'' missing guide structures. 
See the right part of the square pot.
Visualizations by RobSelf-De for real-world NIR SR ($\times2$).
}
\label{supp_fig20_obserations} 
\end{figure}


\subsection{Interesting Observations} 
\label{subsec:interesting}

Compared to off-the-shelf registration modules~\cite{MURF_2023_PAMI},
our translator more effectively handles \textbf{diverse misalignments}, 
\eg, inherent cross-sensor misalignment with large viewpoint variation (\cref{supp_fig13_ga}, left) and non-rigid object motion (\cref{supp_fig13_ga}, right). 
Such robustness stems from its weakly-supervised translation objective, which encourages source-consistent guide features. 

Moreover, it can \textbf{``synthesize'' structures that are missing} from the guide images, thereby still providing effective guidance for source enhancement. 
This arises when some guide structures are absent due to field-of-view differences and occlusions. 
As shown in \cref{supp_fig20_obserations}, our translator recovers in $\mathbf{F}^{\rm Aligned}_{guide}$ the right part of the square pot, which is entirely missing from the original guide images and features. 
With the recovered structures, our filter enhances the source feature and produces promising SR prediction.
A similar phenomenon is observed in the second scene of \cref{fig7:real_rgb_depth}—the left part of the cactus.
We attribute this ability to the translator's tendency to 
``borrow’’ and align contextually related content in the guide, driven by the translation objective toward mimicking the source. 

\begin{figure}[!t]
\centering
\begin{minipage}{0.25\linewidth} 
\includegraphics[width=1\linewidth]{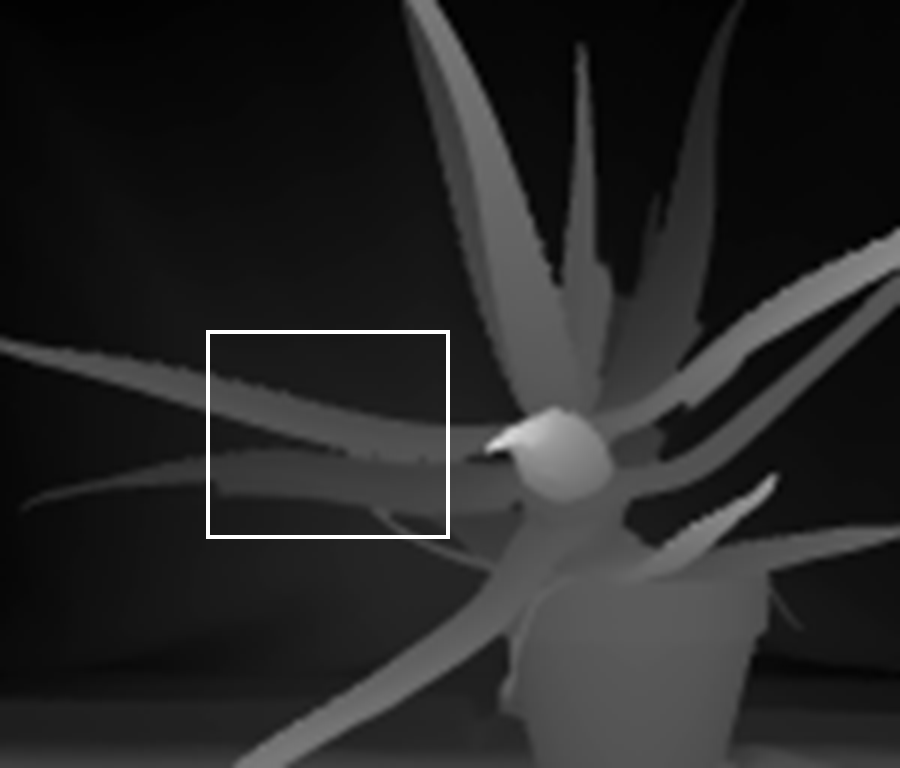}
\end{minipage}
\begin{minipage}{0.25\linewidth} 
\includegraphics[width=1\linewidth]{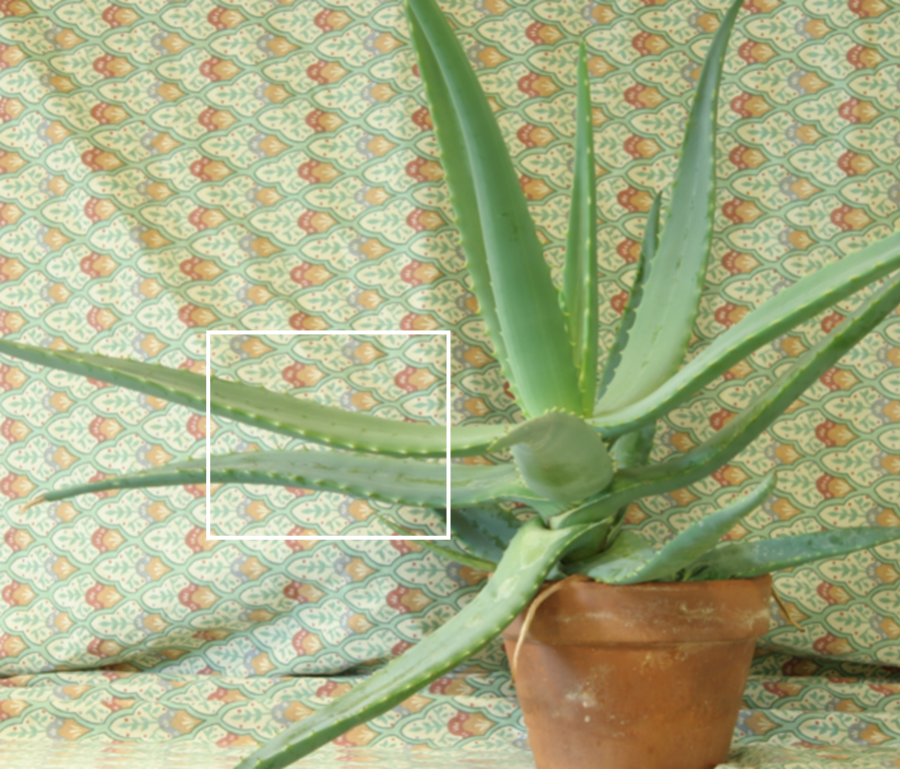}
\end{minipage} 
\begin{minipage}{0.115\linewidth} 
\includegraphics[width=1\linewidth]{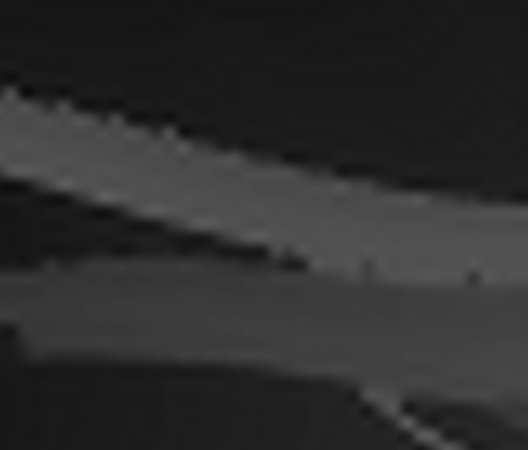}
\subcaption*{\tiny LR Source} 
\includegraphics[width=1\linewidth]{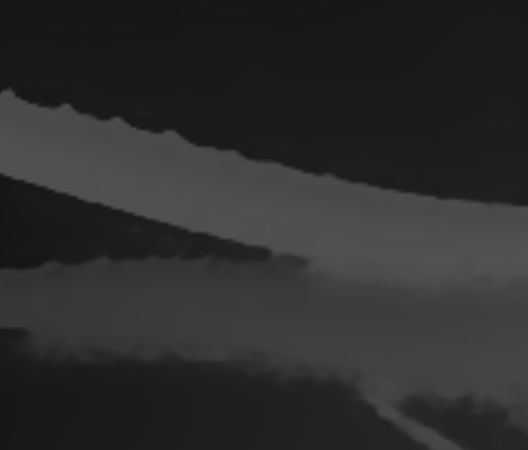}
\subcaption*{\tiny MMSR \cite{MMSR_ECCV_2022}} 
\end{minipage}
\begin{minipage}{0.115\linewidth} 
\includegraphics[width=1\linewidth]{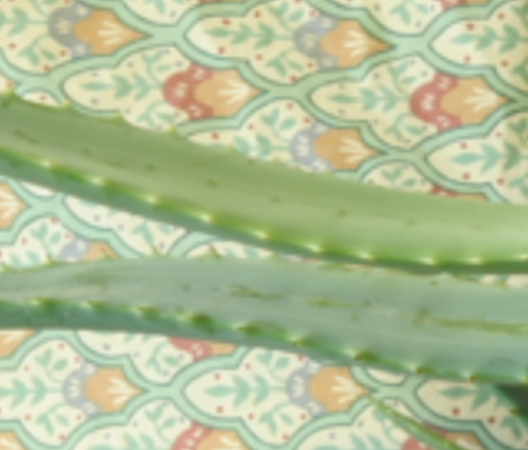}
\subcaption*{\tiny HR Guide} 
\includegraphics[width=1\linewidth]{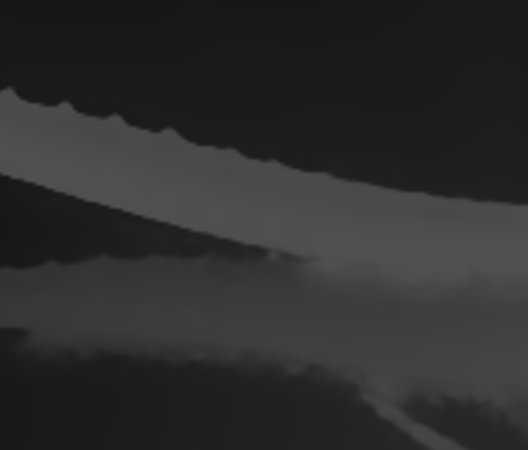}
\subcaption*{\tiny SSGNet \cite{SSGNet_2023_AAAI}} 
\end{minipage}
\begin{minipage}{0.115\linewidth}
\includegraphics[width=1\linewidth]{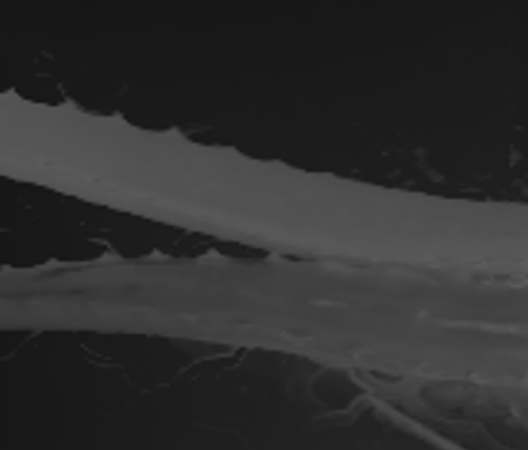}
\subcaption*{\tiny P2P \cite{P2P_2019_ICCV}} 
\includegraphics[width=1\linewidth]{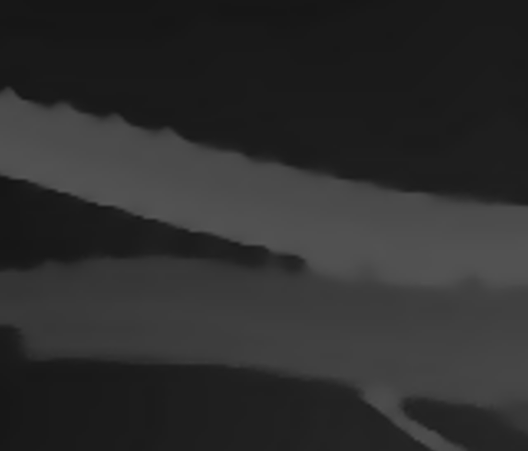}
\subcaption*{\tiny RobSelf-De} 
\end{minipage}
\begin{minipage}{0.115\linewidth} 
\includegraphics[width=1\linewidth]{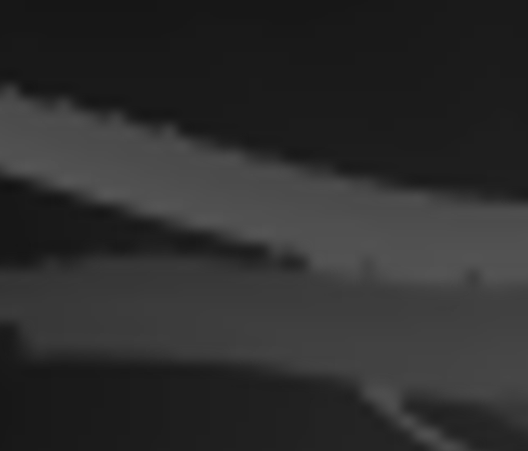}
\subcaption*{\tiny CMSR \cite{CMSR_2021_CVPR}} 
\includegraphics[width=1\linewidth]{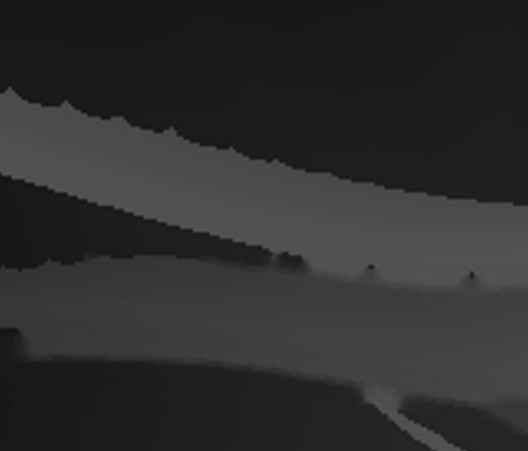}
\subcaption*{\tiny GT} 
\end{minipage} 
\caption{
Performance on well-aligned data: synthesized RGB-guided depth SR ($\times4$).
}
\label{supp_fig22_mis_none} 
\end{figure}

In \textbf{well-aligned scenarios} (\cref{supp_fig22_mis_none}), 
our model achieves competitive performance, 
once again demonstrating that our translator and filter facilitate effective guide utilization and source enhancement.
See the supplementary material for detailed experiments under different misalignment degrees.

\subsection{Runtime Efficiency} 

\Cref{tab5_time} compares the runtime (online optimization + reference) of self-supervised methods on an NVIDIA A100 GPU. 
SSGNet~\cite{SSGNet_2023_AAAI} and MMSR~\cite{MMSR_ECCV_2022} were optimized for 1000 iterations as in our settings. 
P2P~\cite{P2P_2019_ICCV} followed its original settings to ensure performance.
CMSR~\cite{CMSR_2021_CVPR} had to be run on a different GPU due to its TensorFlow 1 code base and is therefore excluded. 
Our models achieve superior efficiency compared to existing self-supervised methods. 
For example, in task \uppercase\expandafter{\romannumeral3}, RobSelf-Re is up to 15.3$\times$ faster than P2P and at least 2.56$\times$ faster than MMSR and SSGNet while achieving lower RMSE. 
Such efficiency stems from our lightweight architecture 
and the filter, which requires no additional processing (\eg, filtering and fusion) for the guide feature. 

\begin{table}[!t]
  \caption{
  Runtime comparison of self-supervised methods. *Pre-alignment applied.
  } 
  \label{tab5_time}
  \centering
  \scalebox{0.7}{ 
  \begin{tabular}{l|ccc|cc}
    \toprule
     & P2P*\cite{P2P_2019_ICCV} & MMSR*\cite{MMSR_ECCV_2022} & SSGNet*\cite{SSGNet_2023_AAAI} & RobSelf-De & RobSelf-Re \\ \hline
    \multicolumn{6}{c}{Task \uppercase\expandafter{\romannumeral1} ($\times4$). Guide: $512\times 576\times 3$; Upsampled source: $512\times 576\times 1$.} \\ \hline
    RMSE$\downarrow$ & 2.33 & 1.88 & 1.92 & \textbf{1.43} & {\ul 1.52} \\ 
    Time & 325s & 137s & 151s & {\ul 100s} & \textbf{71s} \\ \hline
    \multicolumn{6}{c}{Task \uppercase\expandafter{\romannumeral2} ($\times2$). Guide: $576\times 640\times 3$; Upsampled source: $576\times 640\times 1$.} \\ \hline
    RMSE$\downarrow$ & 3.51 & 2.76 & 2.74 & \textbf{2.18} & {\ul 2.23} \\ 
    Time & 982s & 164s & 181s & {\ul 122s} & \textbf{89s} \\ \hline
    \multicolumn{6}{c}{Task \uppercase\expandafter{\romannumeral3} ($\times2$). Guide: $576\times 640\times 3$; Upsampled source: $576\times 640\times 1$.} \\ \hline
    RMSE$\downarrow$ & 18.91 & 8.91 & 9.04 & {\ul 3.12} & \textbf{3.09} \\ 
    Time & 982s & 164s & 181s & {\ul 97s} & \textbf{64s} \\
    \bottomrule
  \end{tabular}
  } 
\end{table}


\section{Conclusions} 

We achieve robust self-supervised cross-modal SR on real-world misaligned data by proposing RobSelf. 
Within RobSelf, a misalignment-aware feature translator reformulates and resolves unsupervised cross-modal and cross-resolution alignment, and a content-aware reference filter enables effective and faithful source enhancement. 
RobSelf demonstrates state-of-the-art performance, robustness, generalizability, and superior efficiency, showing promising applicability to practical scenarios where only unlabeled misaligned data is available.

\section*{Acknowledgements} 

This work was supported by JST FOREST (Grant Number JPMJFR206S), JST NEXUS (Grant Number JPMJNX25CA), and JST CRONOS (Grant Number JPMJCS25K5). 
JL was also supported by the RIKEN Junior Research Associate (JRA) program.


%
%

\begin{thebibliography}{10}
\providecommand{\url}[1]{\texttt{#1}}
\providecommand{\urlprefix}{URL }
\providecommand{\doi}[1]{https://doi.org/#1}

\bibitem{ULB17-VT_2018}
Almasri, F., Debeir, O.: Multimodal sensor fusion in single thermal image super-resolution. arXiv preprint arXiv:1812.09276  (2018)

\bibitem{Trans_2020_CVPR}
Arar, M., Ginger, Y., Danon, D., Bermano, A.H., Cohen-Or, D.: Unsupervised multi-modal image registration via geometry preserving image-to-image translation. In: CVPR (2020)

\bibitem{EPFL_2011_CVPR}
Brown, M., Süsstrunk, S.: Multi-spectral {SIFT} for scene category recognition. In: CVPR (2011)

\bibitem{Recurrent_2023_CVPR}
Cao, S.Y., Zhang, R., Luo, L., Yu, B., Sheng, Z., Li, J., Shen, H.L.: Recurrent homography estimation using homography-guided image warping and focus transformer. In: CVPR (2023)

\bibitem{LuGS++_2026}
Cui, Z., Liu, S., Dong, X., Chu, X., Gu, L., Yang, M.H., Harada, T.: Unifying color and lightness correction with view-adaptive curve adjustment for robust {3D} novel view synthesis. arXiv preprint arXiv:2602.18322  (2026)

\bibitem{DCN_2017_ICCV}
Dai, J., Qi, H., Xiong, Y., Li, Y., Zhang, G., Hu, H., Wei, Y.: Deformable convolutional networks. In: ICCV (2017)

\bibitem{CrossHomo_2024_PAMI}
Deng, X., Liu, E., Gao, C., Li, S., Gu, S., Xu, M.: {CrossHomo}: Cross-modality and cross-resolution homography estimation. IEEE TPAMI  (2024)

\bibitem{PAN_2025_ICCV}
Do, J., Kim, S., Youk, G., Lee, J., Kim, M.: {PAN-Crafter}: Learning modality-consistent alignment for {PAN}-sharpening. In: ICCV (2025)

\bibitem{LowPassDeblur_2023_ICCV}
Dong, J., Pan, J., Yang, Z., Tang, J.: Multi-scale residual low-pass filter network for image deblurring. In: ICCV (2023)

\bibitem{UnalignedHSI_2024_AAAI}
Dong, W., Xu, Y., Qu, J., Hou, S.: Learning multi-modal cross-scale deformable transformer network for unregistered hyperspectral image super-resolution. In: AAAI (2024)

\bibitem{Dong_2024_WACV}
Dong, X., Yokoya, N.: Understanding dark scenes by contrasting multi-modal observations. In: WACV (2024)

\bibitem{MMSR_ECCV_2022}
Dong, X., Yokoya, N., Wang, L., Uezato, T.: Learning mutual modulation for self-supervised cross-modal super-resolution. In: ECCV (2022)

\bibitem{Pansharpening_2024_CVPR}
Duan, Y., Wu, X., Deng, H., Deng, L.J.: Content-adaptive non-local convolution for remote sensing pansharpening. In: CVPR (2024)

\bibitem{InvertTrans_2024_ECCV}
Guo, M.: Unsupervised multi-modal medical image registration via invertible translation. In: ECCV (2024)

\bibitem{UGSR_2021_TIP}
Gupta, H., Mitra, K.: Toward unaligned guided thermal super-resolution. IEEE TIP  (2021)

\bibitem{FDSR_2021_CVPR}
He, L., Zhu, H., Li, F., Bai, H., Cong, R., Zhang, C., Lin, C., Liu, M., Zhao, Y.: Towards fast and accurate real-world depth super-resolution: Benchmark dataset and baseline. In: CVPR (2021)

\bibitem{Hirschmuller_2007_CVPR}
Hirschmuller, H., Scharstein, D.: Evaluation of cost functions for stereo matching. In: CVPR (2007)

\bibitem{DSS_2020_SPL}
Huang, Y., Li, L., Zhu, H., Hu, B.: Blind quality index of depth images based on structural statistics for view synthesis. IEEE SPL  (2020)

\bibitem{CP2D_2025_AAAI}
Kang, J., Cai, Q., Tan, R., Liu, Y., Liu, Z.: {C2PD}: Continuity-constrained pixelwise deformation for guided depth super-resolution. In: AAAI (2025)

\bibitem{DKN_2021_IJCV}
Kim, B., Ponce, J., Ham, B.: Deformable kernel networks for joint image filtering. IJCV  (2021)

\bibitem{PixelNIR_2025_CVPR}
Kim, J., Baek, S.H.: Pixel-aligned {RGB-NIR} stereo imaging and dataset for robot vision. In: CVPR (2025)

\bibitem{Adam_2015_ICLR}
Kingma, D., Ba, J.: Adam: A method for stochastic optimization. In: ICLR (2015)

\bibitem{NYU_2012_ECCV}
Kohli, P., Silberman, N., Hoiem, D., Fergus, R.: Indoor segmentation and support inference from {RGBD} images. In: ECCV (2012)

\bibitem{Indescribable_2023_CVPR}
Kong, L., Qi, X.S., Shen, Q., Wang, J., Zhang, J., Hu, Y., Zhou, Q.: Indescribable multi-modal spatial evaluator. In: CVPR (2023)

\bibitem{UnalignedHSI_2024_TNNLS}
Lai, Z., Fu, Y., Zhang, J.: Hyperspectral image super resolution with real unaligned {RGB} guidance. IEEE TNNLS  (2024)

\bibitem{MulFS-CAP_2025_PAMI}
Li, H., Yang, Z., Zhang, Y., Jia, W., Yu, Z., Liu, Y.: {MulFS-CAP}: Multimodal fusion-supervised cross-modality alignment perception for unregistered infrared-visible image fusion. IEEE TPAMI  (2025)

\bibitem{ToFGaussian_2025_CVPR}
Li, R., Okunev, M., Guo, Z., Duong, A.H., Richardt, C., O'Toole, M., Tompkin, J.: Time of the flight of the gaussians: Optimizing depth indirectly in dynamic radiance fields. In: CVPR (2025)

\bibitem{Studio_2025_CVPR}
Lincetto, F., Agresti, G., Rossi, M., Zanuttigh, P.: Multimodalstudio: A heterogeneous sensor dataset and framework for neural rendering across multiple imaging modalities. In: CVPR (2025)

\bibitem{NLRN_2018_NIPS}
Liu, D., Wen, B., Fan, Y., Loy, C.C., Huang, T.S.: Non-local recurrent network for image restoration. In: NeurIPS (2018)

\bibitem{P2P_2019_ICCV}
Lutio, R.d., D'Aronco, S., Wegner, J.D., Schindler, K.: Guided super-resolution as pixel-to-pixel transformation. In: ICCV (2019)

\bibitem{Mei_2021_CVPR}
Mei, Y., Fan, Y., Zhou, Y.: Image super-resolution with non-local sparse attention. In: CVPR (2021)

\bibitem{DADA_2023_CVPR}
Metzger, N., Daudt, R.C., Schindler, K.: Guided depth super-resolution by deep anisotropic diffusion. In: CVPR (2023)

\bibitem{AgnosticMedical_2024_CVPR}
Mok, T.C.W., Li, Z., Bai, Y., Zhang, J., Liu, W., Zhou, Y.J., Yan, K., Jin, D., Shi, Y., Yin, X., Lu, L., Zhang, L.: Modality-agnostic structural image representation learning for deformable multi-modality medical image registration. In: CVPR (2024)

\bibitem{Refine_2019_ICCV}
Qiu, D., Pang, J., Sun, W., Yang, C.: Deep end-to-end alignment and refinement for time-of-flight {RGB-D} module. In: ICCV (2019)

\bibitem{Fusion_2025_TIP}
Qu, J., Wu, X., Dong, W., Cui, J., Li, Y.: {IR\&ArF}: Toward deep interpretable arbitrary resolution fusion of unregistered hyperspectral and multispectral images. IEEE TIP  (2025)

\bibitem{MINIMA_2025_CVPR}
Ren, J., Jiang, X., Li, Z., Liang, D., Zhou, X., Bai, X.: {MINIMA}: Modality invariant image matching. In: CVPR (2025)

\bibitem{Scharstein_2007_CVPR}
Scharstein, D., Pal, C.: Learning conditional random fields for stereo. In: CVPR (2007)

\bibitem{CMSR_2021_CVPR}
Shacht, G., Danon, D., Fogel, S., Cohen-Or, D.: Single pair cross-modality super resolution. In: CVPR (2021)

\bibitem{RS_2023_TGRS}
Shi, L., Zhao, R., Pan, B., Zou, Z., Shi, Z.: Unsupervised multimodal remote sensing image registration via domain adaptation. IEEE TGRS  (2023)

\bibitem{SSGNet_2023_AAAI}
Shin, J., Shin, S., Jeon, H.G.: Task-specific scene structure representations. In: AAAI (2023)

\bibitem{LTFv2_2020_NIPS}
Song, L., Li, Y., Jiang, Z., Li, Z., Zhang, X., Sun, H., Sun, J., Zheng, N.: Rethinking learnable tree filter for generic feature transform. In: NeurIPS (2020)

\bibitem{LTF_2019_NIPS}
Song, L., Li, Y., Li, Z., Yu, G., Sun, H., Sun, J., Zheng, N.: Learnable tree filter for structure-preserving feature transform. In: NeurIPS (2019)

\bibitem{Alternating_2024_NIPS}
Song, S., Lew, J., Jang, H., Yoon, S.: Unsupervised homography estimation on multimodal image pair via alternating optimization. In: NeurIPS (2024)

\bibitem{NLCRF_2022_CVPR}
Veksler, O., Boykov, Y.: Sparse non-local {CRF}. In: CVPR (2022)

\bibitem{NL_2018_CVPR}
Wang, X., Girshick, R., Gupta, A., He, K.: Non-local neural networks. In: CVPR (2018)

\bibitem{Video_2026_AAAI}
Wang, Z., Wu, Y., Li, X., Yan, Z., Yang, J.: Spatiotemporal difference network for video depth super-resolution. In: AAAI (2026)

\bibitem{DORNet_2025_CVPR}
Wang, Z., Yan, Z., Pan, J., Gao, G., Zhang, K., Yang, J.: {DORNet}: A degradation oriented and regularized network for blind depth super-resolution. In: CVPR (2025)

\bibitem{MOMNet_2026}
Wang, Z., Yan, Z., Wu, Y., Gao, G., Li, X., Yang, J.: Multi-order matching network for alignment-free depth super-resolution. arXiv preprint arXiv:2511.16361  (2026)

\bibitem{SGNet_2024_AAAI}
Wang, Z., Yan, Z., Yang, J.: {SGNet}: Structure guided network via gradient-frequency awareness for depth map super-resolution. In: AAAI (2024)

\bibitem{SPFNet_2026_IJCV}
Wang, Z., Yan, Z., Yang, M.H., Pan, J., Yang, J., Tai, Y., Gao, G.: Scene prior filtering for depth map super-resolution. IJCV  (2026)

\bibitem{DEPTHOR_2025_ICCV}
Xiang, J., Zhu, X., Wang, X., Wang, Y., Zhang, H., Guo, F., Yang, X.: {DEPTHOR}: Depth enhancement from a practical light-weight {dToF} sensor and {RGB} image. In: ICCV (2025)

\bibitem{MURF_2023_PAMI}
Xu, H., Yuan, J., Ma, J.: {MURF}: Mutually reinforcing multi-modal image registration and fusion. IEEE TPAMI  (2023)

\bibitem{Reconstruct_2022_TMI}
Xuan, K., Xiang, L., Huang, X., Zhang, L., Liao, S., Shen, D., Wang, Q.: Multi-modal {MRI} reconstruction assisted with spatial alignment network. IEEE TMI  (2022)

\bibitem{Yan_2022_TNNLS}
Yan, Z., Wang, K., Li, X., Zhang, Z., Li, G., Li, J., Yang, J.: Learning complementary correlations for depth super-resolution with incomplete data in real world. IEEE TNNLS  (2022)

\bibitem{DuCos_2025_ICCV}
Yan, Z., Wang, Z., Dong, H., Li, J., Yang, J., Lee, G.H.: Ducos: Duality constrained depth super-resolution via foundation model. In: ICCV (2025)

\bibitem{DegBins_2026}
Yan, Z., Wang, Z., Yang, J., Lee, G.H.: {DegBins}: Degradation-driven binning for depth super-resolution. arXiv preprint arXiv:2605.09628  (2026)

\bibitem{OmniSegment_2025_NIPS}
Yin, B.W., Cao, J.L., Zhang, X., Chen, Y., Cheng, M.M., Hou, Q.: {OmniSegmentor}: A flexible multi-modal learning framework for semantic segmentation. In: NeurIPS (2025)

\bibitem{SFG_2023_AAAI}
Yuan, J., Jiang, H., Li, X., Qian, J., Li, J., Yang, J.: Structure flow-guided network for real depth super-resolution. In: AAAI (2023)

\bibitem{SDME_2024_ICML}
Zhang, K., Ma, J.: Sparse-to-dense multimodal image registration via multi-task learning. In: ICML (2024)

\bibitem{NIQE_2015_TIP}
Zhang, L., Zhang, L., Bovik, A.C.: A feature-enriched completely blind image quality evaluator. IEEE TIP  (2015)

\bibitem{SCPNet_2024_ECCV}
Zhang, R., Ma, J., Cao, S.Y., Luo, L., Yu, B., Chen, S.J., Li, J., Shen, H.L.: {SCPNet}: Unsupervised cross-modal homography estimation via intra-modal self-supervised learning. In: ECCV (2024)

\bibitem{CDFDSR_2025_IF}
Zhang, S., Dong, J., Ma, Y., Cai, H., Wang, M., Li, Y., Kabika, T.B., Li, X., Hou, W.: {CDF-DSR}: Learning continuous depth field for self-supervised {RGB}-guided depth map super resolution. Information Fusion  (2025)

\bibitem{UnalignedHSI_2025_IJCV}
Zhang, Y., Lai, Z., Zhang, T., Fu, Y., Zhou, C.: Unaligned {RGB} guided hyperspectral image super-resolution with spatial-spectral concordance. IJCV  (2025)

\bibitem{Lucas_2021_CVPR}
Zhao, Y., Huang, X., Zhang, Z.: Deep {Lucas-Kanade} homography for multimodal image alignment. In: CVPR (2021)

\bibitem{Video_2025_NIPS}
Zhao, Z., Bai, H., Ke, B., Cui, Y., Deng, L., Zhang, Y., Zhang, K., Schindler, K.: A unified solution to video fusion: From multi-frame learning to benchmarking. In: NeurIPS (2025)

\bibitem{Equi_2024_CVPR}
Zhao, Z., Bai, H., Zhang, J., Zhang, Y., Zhang, K., Xu, S., Chen, D., Timofte, R., Van~Gool, L.: Equivariant multi-modality image fusion. In: CVPR (2024)

\bibitem{DDFM_2023_ICCV}
Zhao, Z., Bai, H., Zhu, Y., Zhang, J., Xu, S., Zhang, Y., Zhang, K., Meng, D., Timofte, R., Van~Gool, L.: {DDFM}: Denoising diffusion model for multi-modality image fusion. In: ICCV (2023)

\bibitem{Spherical_2023_ICCV}
Zhao, Z., Zhang, J., Gu, X., Tan, C., Xu, S., Zhang, Y., Timofte, R., Van~Gool, L.: Spherical space feature decomposition for guided depth map super-resolution. In: ICCV (2023)

\bibitem{DCTNet_2022_CVPR}
Zhao, Z., Zhang, J., Xu, S., Lin, Z., Pfister, H.: Discrete cosine transform network for guided depth map super-resolution. In: CVPR (2022)

\bibitem{Bandpass_2020_CVPR}
Zheng, B., Yuan, S., Slabaugh, G., Leonardis, A.: Image demoireing with learnable bandpass filters. In: CVPR (2020)

\bibitem{GraphUnfold_2026_CVPR}
Zhong, Z., Chen, P., Shen, Q., Li, B., Wang, S.: Dual graph regularized deep unfolding network for guided depth map super-resolution. In: CVPR (2026)

\bibitem{DCNAS_2025_TPAMI}
Zhong, Z., Liu, X., Jiang, J., Zhao, D., Wang, S.: Dual-level cross-modality neural architecture search for guided image super-resolution. IEEE TPAMI  (2025)

\end{thebibliography}

\end{document}